\documentclass[lettersize,journal]{IEEEtran}
\usepackage{amsmath,amssymb}
\usepackage{psfrag}
\usepackage{epsfig}
\usepackage{cite}
\usepackage{graphics}
\usepackage{color}
\usepackage{cases}
\usepackage{subfigure}
\usepackage{epstopdf}
\usepackage{epsfig}
\usepackage{hyperref}
\usepackage{multirow}
\usepackage{mathrsfs}

\usepackage{algorithm}
\usepackage{algpseudocode}
\usepackage{url}

\usepackage{booktabs}
\usepackage{pifont}

\usepackage{paracol}
\usepackage[table]{xcolor}
\usepackage{colortbl}
\renewcommand{\vec}[1]{\boldsymbol{#1}}
\newcommand{\cmark}{\ding{51}}%
\newcommand{\xmark}{\ding{55}}%
\algnewcommand{\IfThenElse}[3]{
  \State \algorithmicif\ #1\ \algorithmicthen\ #2\ \algorithmicelse\ #3}

\begin{document}

\title{Improving Fast Adversarial Training via Self-Knowledge Guidance}


\author{ 
    Chengze Jiang, Junkai Wang, Minjing Dong, Jie Gui,~\IEEEmembership{Senior Member,~IEEE,} Xinli Shi,~\IEEEmembership{Senior Member,~IEEE,} Yuan Cao, Yuan Yan Tang,~\IEEEmembership{Life Fellow,~IEEE,} James Tin-Yau Kwok,~\IEEEmembership{Fellow,~IEEE}

\thanks{C. Jiang, J. Wang, and X. Shi are with the School of Cyber Science and Engineering, Southeast University, Nanjing 210000, China (e-mail: czjiang@seu.edu.cn; 220224901@seu.edu.cn; xinli$\_$shi@seu.edu.cn).}
\thanks{M. Dong is with the Department of Computer Science, City University of Hong Kong. (e-mail: minjdong@cityu.edu.hk).}
\thanks{J. Gui is with the School of Cyber Science and Engineering, Southeast University and with Purple Mountain Laboratories, Nanjing 210000, China (e-mail: guijie@seu.edu.cn).}
\thanks{Y. Cao is with the School of Information Science and Engineering, Ocean University of China, Qingdao 266100, China, (e-mail: cy8661@ouc.edu.cn).}
\thanks{Y. Tang is with the Department of Computer and Information Science, University of Macau, Macau 999078, China (e-mail: yytang@um.edu.mo).}
\thanks{J. T. -Y. Kwok is with the Department of Computer Science and Engineering, The Hong Kong University of Science and Technology, Hong Kong, China (e-mail: jamesk@cse.ust.hk).}
}

\markboth{Manuscript for IEEE Transactions on Information Forensics and Security}%
{Shell \MakeLowercase{\textit{et al.}}: A Sample Article Using IEEEtran.cls for IEEE Journals}

\maketitle

\begin{abstract}
Adversarial training has achieved remarkable advancements in defending against adversarial attacks. Among them, fast adversarial training (FAT) is gaining attention for its ability to achieve competitive robustness with fewer computing resources. Existing FAT methods typically employ a uniform strategy that optimizes all training data equally without considering the influence of different examples, which leads to an imbalanced optimization. However, this imbalance remains unexplored in the field of FAT. In this paper, we conduct a comprehensive study of the imbalance issue in FAT and observe an obvious class disparity regarding their performances. This disparity could be embodied from a perspective of alignment between clean and robust accuracy. Based on the analysis, we mainly attribute the observed misalignment and disparity to the imbalanced optimization in FAT, which motivates us to optimize different training data adaptively to enhance robustness. Specifically, we take disparity and misalignment into consideration. First, we introduce self-knowledge guided regularization, which assigns differentiated regularization weights to each class based on its training state, alleviating class disparity. Additionally, we propose self-knowledge guided label relaxation, which adjusts label relaxation according to the training accuracy, alleviating the misalignment and improving robustness. By combining these methods, we formulate the Self-Knowledge Guided FAT (SKG-FAT), leveraging naturally generated knowledge during training to enhance the adversarial robustness without compromising training efficiency. Extensive experiments on four standard datasets demonstrate that the SKG-FAT improves the robustness and preserves competitive clean accuracy, outperforming the state-of-the-art methods.
\end{abstract}


\section{Introduction}
Deep learning models have demonstrated impressive performance across various applications \cite{han2022survey, mao2022towards}. However, they remain vulnerable to adversarial attacks, which raise significant security threats \cite{RLFA, FASTEN, chen2023adaptive}. Adversarial attacks craft elaborate perturbations to clean examples which can deceive the model into making incorrect predictions \cite{10504304}. In response, considerable research is devoted to developing defense methods \cite{AADGD, AEANRF, huang2024pointcat}. Among them, adversarial training emerges as a promising approach to enhance the robustness of deep learning models \cite{FATASS, FSRAR}. Adversarial training involves augmenting the training dataset with adversarial examples (AEs) to enhance the adversarial robustness of the model \cite{TopAlignAT, MutAT, 10189878}. However, multi-step adversarial training utilizes the projected gradient descent (PGD) for generating AEs \cite{PGD}, thereby requiring multiple extra gradient backpropagations, which is computationally expensive \cite{singh2024revisiting}. To achieve efficient adversarial training, fast adversarial training (FAT) adopts the fast gradient sign method (FGSM) to replace PGD \cite{FGSMRS, NFGSM}. This substitution enables adversarial training to require only one additional backpropagation for generating AEs compared to normal training \cite{RFATLAP}, leading to a notable reduction in time consumption \cite{SAT}.
\par
Recent advancements in adversarial training primarily focus on enhancing both efficiency and adversarial robustness \cite{RAFATLB, SADR}. These variants typically employ a uniform training strategy across all examples, disregarding the accuracy diversity arising from example feature differences \cite{yue2024revisiting}. In standard adversarial training, it has been reported that disparities in robust accuracy among examples result in significant imbalances in robustness across different classes \cite{TBRTBF, CFA}. Consequently, several studies seek to address this issue within the multi-step adversarial training paradigm by exploiting and mitigating these imbalances to improve adversarial robustness \cite{CFA, DAFA}. However, the issue of imbalance in FAT lacks exploration, which naturally raises a question: \textit{Does a similar issue also exist in FAT, and how does it manifest?} Moreover, whether can we develop new methods to improve the adversarial robustness by mitigating this issue is also undetermined. 
\par 
Motivated by these observations, we propose to investigate the potential imbalance issue in the field of FAT from empirical evidence and corresponding analysis, which guides the development of effective FAT with superior adversarial robustness and computational efficiency. Specifically, we first conduct a comprehensive empirical analysis from the class-wise perspective within the FAT paradigm. Our results reveal substantial differences in robust accuracy across different classes. Besides, we also discover that examples from partial classes exhibit counter-intuitive observations that there exists a misalignment between clean and robust accuracy, uncovering intriguing characteristics from a different perspective. To further explore the issues behind these phenomena, we propose to analyze the impact of diverse examples by categorizing examples into four groups based on the accuracy alignment, where each group contains training data with similar classification performance. Based on the analysis, we mainly attribute these issues to the fact that all the training data are treated equally in FAT. Thus, we are inspired to explore methods to enhance FAT performance by optimizing training data adaptively. Consequently, we propose self-knowledge guided FAT (SKG-FAT), which incorporates self-knowledge guided regularization and label relaxation. The regularization allocates differentiated regularization weights to each class based on its training state and aligns accuracy, thereby relieving accuracy disparity. Meanwhile, the label relaxation dynamically adjusts the label relaxation according to the training performance of each class, improving training stability and overall model robustness. Extensive experiments demonstrate that our SKG-FAT framework significantly improves robust accuracy across various attacks. The main contributions are highlighted as
\begin{itemize}
	\item We investigate and analyze the imbalanced optimization within the FAT from a class-wise perspective. Besides, a perspective of accuracy alignment is introduced to conduct a deeper analysis of robust disparities as a whole.
	\item Motivated by our observations, the SKG-FAT is proposed, which incorporates self-knowledge guided regularization and label relaxation. Our SKG-FAT can enhance adversarial robustness without compromising training efficiency.
	\item Comprehensive experiments on four benchmark datasets are performed to evaluate the proposed SKG-FAT. Results demonstrate that our SKG-FAT effectively enhances the robustness of models across different datasets.
\end{itemize}
\par
The rest of this paper is organized into five sections. The preliminaries and related works are provided in Section \ref{PRWs}. The problem analysis from perspectives of class and accuracy alignment under FAT are presented in Section \ref{PAs}. Section \ref{Met} presents the proposed methodology and implementation details, aided by pseudocode. Subsequently, comprehensive experiments are conducted to verify the effectiveness of the proposed SKG-FAT in Section \ref{EASec}. Finally, the conclusion and future research directions are given in Section \ref{CFSec}.

\section{Preliminaries and Related Works}\label{PRWs}
\subsection{Preliminaries and Notations}
Define a classifier $f_{\theta}(\vec{x}): \vec{x}\to \vec{y}$ with weight parameter $\theta$ for the classification task of dataset $\mathcal{X}=\{\vec{x}_1,\cdots, \vec{x}_n\}$ with label $\mathcal{Y}$ subject to distribution $\mathcal{D}$. For example $\vec{x}\in\mathcal{X}$, the groundtruth label is represented as $\vec{y}\in\mathbb{R}^{m}$, where $n$ and $m$ denotes the number of examples and classes in the dataset, respectively. Then, the cross-entropy loss $\mathcal{L}(f_{\theta}(\vec{x}),\vec{y})$ is adopted to scale the performance of the classifier $f_{\theta}$, and the empirical risk of the classifier $f_{\theta}$ is formulated as $\mathcal{L}(f_{\theta}(\mathcal{X}),\mathcal{Y})=\mathbb{E}_{\vec{x}\in\mathcal{X}}[\mathcal(f_\theta(\vec{x}), \vec{y})]$ \cite{CFA}.
\subsection{Multi-Step Adversarial Training}
Adversarial training improves the robustness of the model $f_{\theta}(\cdot)$ by adding perturbations to clean examples, thereby generating AEs $\vec{x}'$, which are then incorporated into the training data \cite{ATPD, LBGAT, wang2023better}. This process can be mathematically represented as a minimax optimization problem:
\begin{equation}\label{ATDefine}
	\min_{\theta} \Big{\{}\mathbb{E}_{\{\vec{x}, \vec{y}\}\sim\mathcal{D}}\Big{[}\max_{\vec{x}':\|\vec{x}'-\vec{x}\|_{\text{p}}\leq \epsilon} \mathcal{L}\big{(}f_{\theta}(\vec{x}'), \vec{y}\big{)}\Big{]}\Big{\}},
\end{equation}
where $\epsilon$ denotes the perturbation budget, $\|\cdot\|_{\text{p}}$ is the $p$-norm operator, and AE $\vec{x}'=\vec{x}+\vec{\delta}$ with $\vec{\delta}$ representing the adversarial perturbation \cite{PGD, BATHE}. The internal optimization problem towards maximizing the classification loss is achieved by generating the worst-case AEs. Meanwhile, the model requires correctly classifying AEs to minimize the classification loss \cite{FATSC, huang2024accelerated}. Multi-step adversarial training exploits PGD for generating AEs to achieve internal maximization within the paradigm \eqref{ATDefine}. The iterative role of PGD is defined as follows:
\begin{equation}
	\vec{x}^{'}_{t+1}=\Phi_{\epsilon}\Big{(}\vec{x}^{'}_{t}+\alpha\cdot\text{sign}\big{(}\nabla_{\vec{x}^{'}_{t}}\mathcal{L}(f_{\theta}(\vec{x}^{'}_{t}), \vec{y})\big{)}\Big{)},
\end{equation}
where $\Phi_{\epsilon}(\cdot)$ denotes the projection operator, $\text{sign}(\cdot)$ represents the sign function, $\alpha$ is step size, and $\nabla_{\vec{x}}$ denotes the gradient of loss $\mathcal{L}(f_{\theta}(\vec{x}), \vec{y})$ respect to $\vec{x}$. Multi-step adversarial training executes multiple backward propagations during each training step, depending on the iteration number of PGD \cite{TEEAT}. Although this approach improves model robustness, it significantly increases the training time, requiring more than several or even ten times the normal training duration. \cite{FGSMRS}.

\subsection{Fast Adversarial Training}
To enhance model robustness while minimizing the training time expenses, FAT employs FGSM instead of PGD to generate AEs for training. FGSM requires only a single backward propagation step to produce AEs \cite{jia2024revisiting, UCOSS}, significantly reducing training time compared to multi-step adversarial training methods \cite{GradAlign, EATTAE}. The specific definition of FGSM within FAT is introduced as follows:
\begin{equation}\label{FGSMFAT}
	\vec{x}' = \vec{x}+ \Phi_{\epsilon}\Big{(}\vec{\delta}_0 + \epsilon \cdot \text{sign}\big{(}\nabla_{\vec{x}}\mathcal{L}(f_{\theta}(\vec{x}+\vec{\delta}_0), \vec{y})\big{)}\Big{)},
\end{equation}
where $\vec{\delta}_0$ represents the initialization perturbation \cite{NFGSM}. For the case of $\Phi_{\epsilon}(\cdot)$ being identity map and $\vec{\delta}_0$ equals zero, the equation \eqref{FGSMFAT} is the formula of FGSM \cite{FGSM}. To improve the diversity of AEs and robustness, FGSM-RS samples from a uniform distribution $\mathcal{U}(-\epsilon, \epsilon)$ to generate initialization perturbation $\vec{\delta}_0$ \cite{FGSMRS}. This method can effectively alleviate catastrophic overfitting and improve robustness. On this basis, the Noise-FGSM (NFGSM) removes the projection operator $\Phi_{\epsilon}(\cdot)$ and increases noise intensity to enhance the training stability \cite{NFGSM}. After that, the sample-dependent adversarial initialization FGSM (FGSM-SDI) is introduced to enhance the quality of AEs \cite{SDI}. This approach leverages a generator to improve AEs quality, thereby boosting overall performance. Besides, GradAlign introduces a regularization that aligns the gradients between clean examples and AEs as $\mathbb{E}_{\vec{x}\sim\mathcal{D}}[\cos (\nabla_{\vec{x}}\mathcal{L}(f_{\theta}(\vec{x}), \vec{y}), \nabla_{\vec{x}'}\mathcal{L}(f_{\theta}(\vec{x}'), \vec{y}))]$, where $\cos (\cdot)$ denotes cosine similarity \cite{GradAlign}. GradAlign mitigates catastrophic overfitting and even makes the performance of FAT more competitive than multi-step adversarial training \cite{TEEAT}. Subsequently, Jia $et~al.$ present the FGSM-PGI \cite{PGIFGSM}, which leverages historical adversarial perturbations obtained from the last batch (PGI-BP), last epoch (PGI-EP), previous all epoch (PGI-MEP) as the initialization for generating adversarial perturbations. FGSM-PGI also contains a regularization to minimize the prediction gap between clean examples and AEs. This method significantly enhances the performance of FAT. Although increasing the perturbation budget can improve model robustness, the above methods suffer unavoidable catastrophic overfitting when training with larger perturbation budgets \cite{NFGSM}. Therefore, Zhao $et~al.$ present FAT with smooth convergence (FGSM-SC) for a smooth loss convergence that eliminates catastrophic overfitting \cite{FATSC}. 
\subsection{Limitation of Existing Solutions and Overall Motivations}
While existing methods have attempted to enhance FAT by addressing various aspects such as initialization, regularization, and mitigating catastrophic overfitting, they adopt a unified optimization and processing for all examples, overlooking the impact of robust accuracy disparity among examples \cite{TBRTBF, DAFA}. The performance variations caused by accuracy disparity in standard adversarial training \cite{WAT} prompt us to investigate whether FAT exhibits similar characteristics. Meanwhile, recent advances demonstrate that addressing or exploiting these performance disparities can significantly improve adversarial robustness \cite{CFA}. This drives our exploration of performance disparities within the FAT paradigm, leveraging empirical analysis to guide novel strategies that enhance FAT performance without extensive hyperparameter tuning or sacrificing computational efficiency. To this end, we design and conduct comprehensive experiments, analyzing FAT from a class-wise perspective and uncovering some counter-intuitive phenomena. As a result, we introduce a new analytical framework that examines the holistic properties of examples. Building on the insights gained from our empirical analysis, we toward to develop a self-knowledge guided method that enhances adversarial robustness while maintaining high computational efficiency and avoiding redundant hyperparameter tuning.

\section{Problem Analysis}\label{PAs}
\subsection{Analysis from Class-Wise Perspective}\label{FCWP}
To explore the training accuracy disparities and their manifestation under the FAT paradigm, we reexamine the class-wise clean and robust accuracy. We conduct experiments using ResNet-18 on the CIFAR-10 dataset with two representative FAT methods: FGSM-RS and PGI-BP. The perturbation budget epsilon is set to 8/255 and step size alpha is set to 8/255. Given that the vanilla FGSM-RS is prone to catastrophic overfitting, we introduced a regularization technique to align the outputs of clean examples and AEs \cite{GAT}. This regularization helps prevent catastrophic overfitting, enabling a more reliable analysis. Figure \ref{CWAcc} presents the class-wise clean and robust accuracy of ResNet18 on the CIFAR-10 training set using FGSM-RS \cite{FGSMRS} and PGI-BP \cite{PGIFGSM}. As illustrated, there are notable differences in the robust and clean accuracy across different classes. Additionally, Figure \ref{CWAcc} reveals that the MSE loss also varies among classes, and identifies a pattern consistent across different FAT methods. Note that while such investigations have been reported in prior studies on standard adversarial training, we extend these findings to FAT, confirming their presence in this context as well. Moreover, our analysis uncovers deeper and more intriguing phenomena that have not been previously reported. Specifically, the top three classes with the highest clean accuracy consist of features that are easily learned by neural networks, leading to superior robust accuracy. Conversely, the bottom two classes with the lowest clean accuracy demonstrate the opposite trend. In other words, improving robust accuracy is generally easier in classes where clean accuracy is more readily improved. According to the trends depicted in Fig.\ref{CWAcc}(a) and (b) for different class curves, it appears that there exists a positive correlation between clean and robust accuracy in each class. This prompts us to raise an intuitive question: 
\begin{figure}[t]\centering
	\includegraphics[scale=0.26]{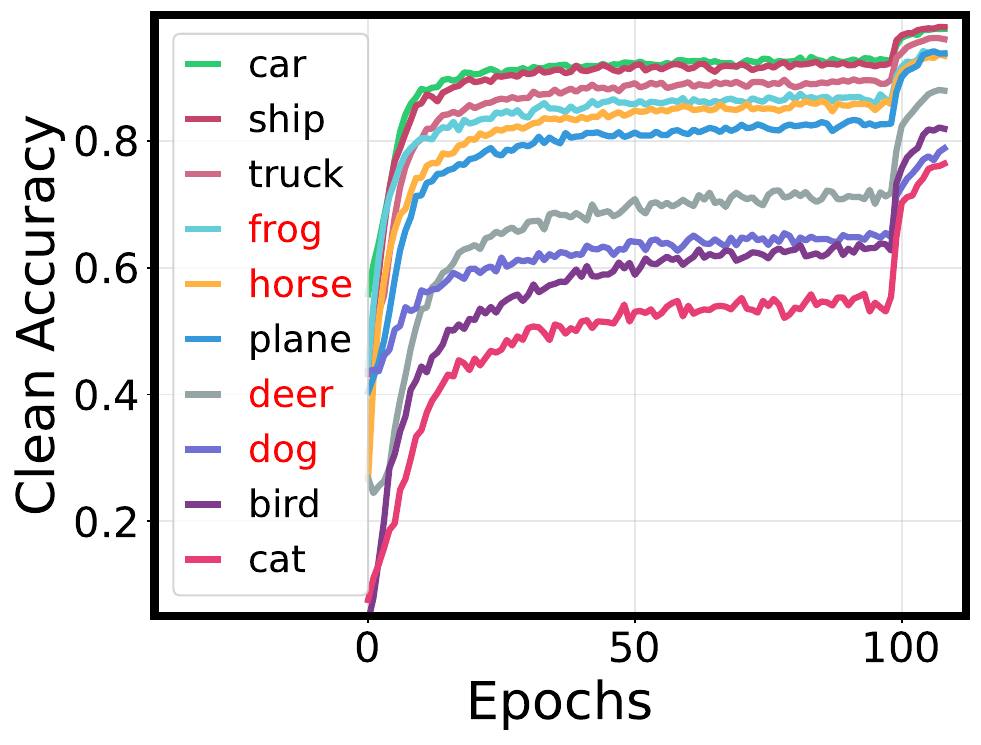}
	\includegraphics[scale=0.26]{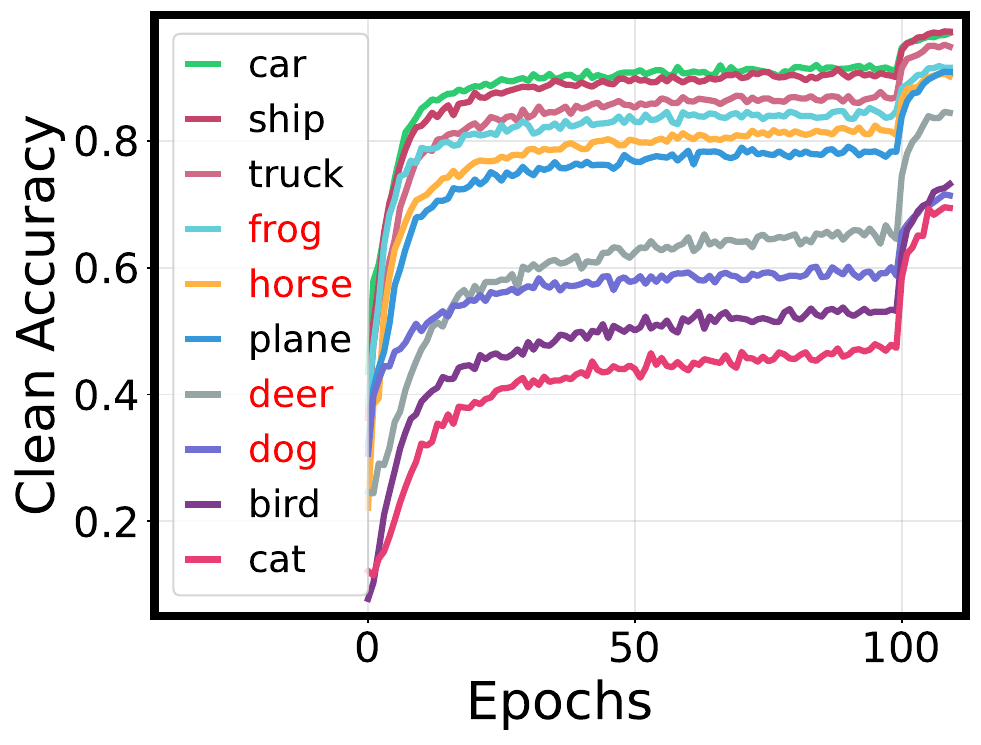}
	\includegraphics[scale=0.26]{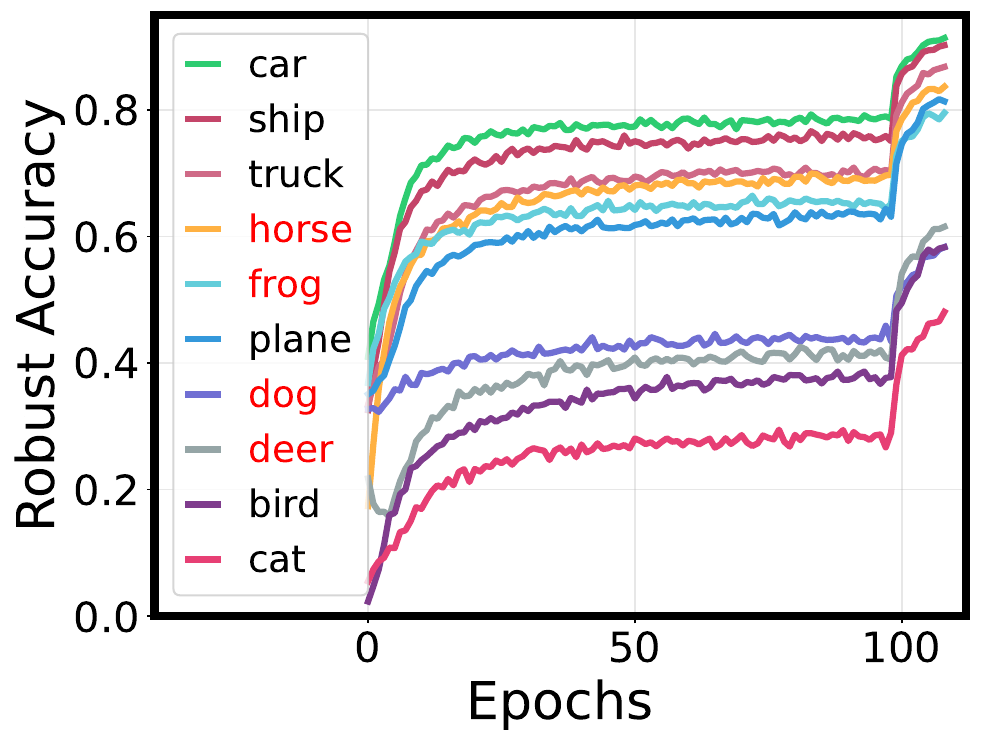}
	\includegraphics[scale=0.26]{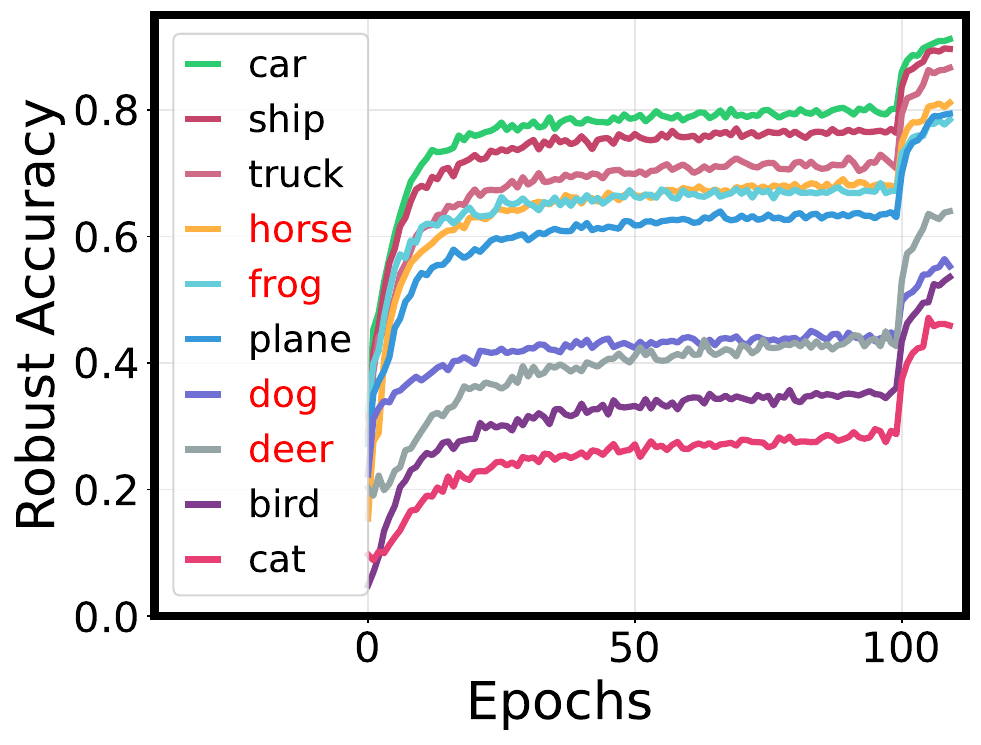}
	\subfigure[]{\includegraphics[scale=0.26]{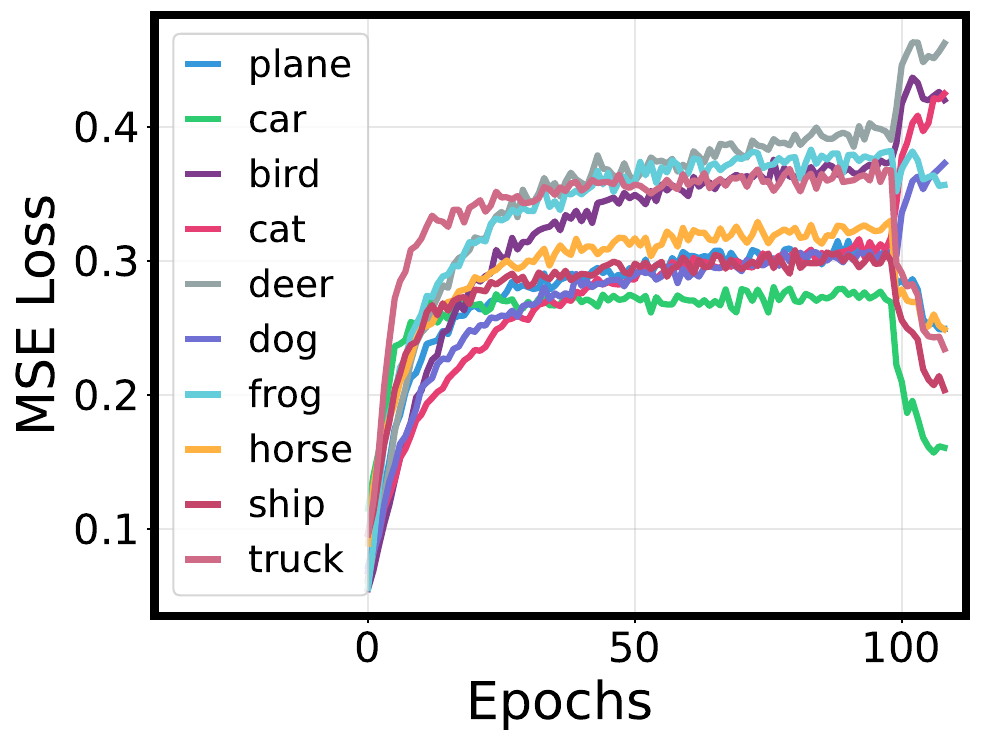}}
	\subfigure[]{\includegraphics[scale=0.26]{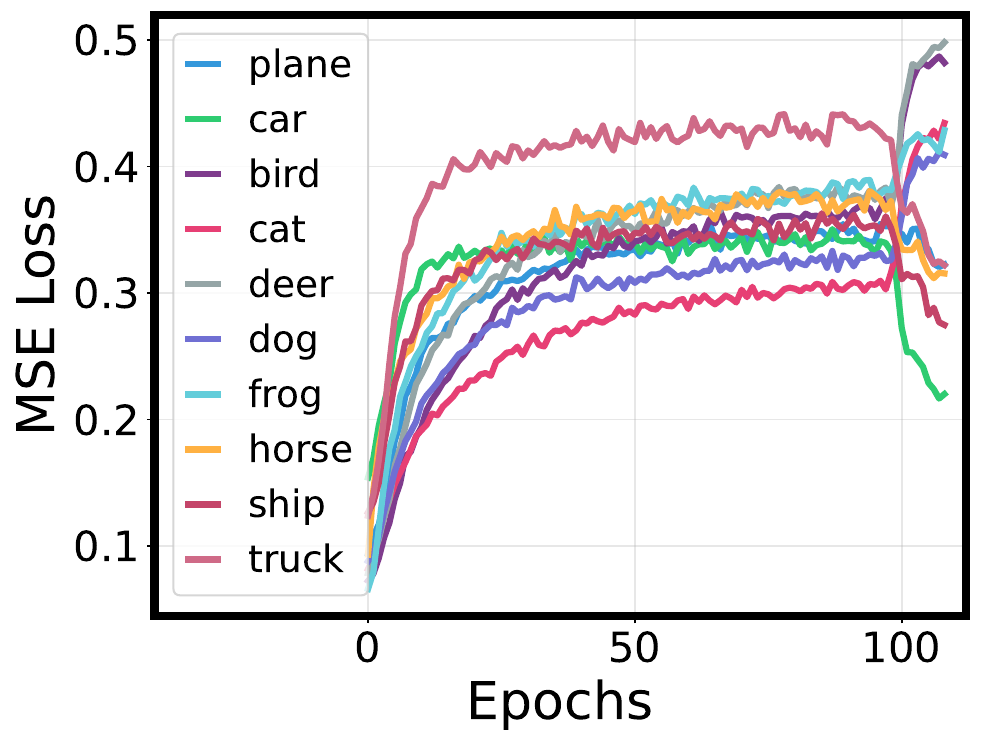}}
	\caption{Class-wise training results on CIFAR-10 training set. (a) Results obtained by FGSM-RS. (b) Results obtained by PGI-BP.}
	\label{CWAcc}
\end{figure}
\textit{Can we develop differentiated training strategies for different classes, thereby improving adversarial robustness?} Inspired by the above observations and this question, we propose two methods in subsections \ref{CWER} and \ref{LERDef} to enhance FAT.
\par
Meanwhile, the positive correlation between clean and robust accuracy weakens for classes with middle-range accuracy, such as horse' and frog' or dog' and deer'. The competitive clean accuracy of these classes does not necessarily translate to robust accuracy, resulting in a misalignment between their rankings in clean and robust accuracy (as highlighted in red). We identify this issue as being caused by the misalignment between clean and robust performance. This counter-intuitive disparity cannot be fully analyzed from a purely class-wise perspective. Therefore, we present a perspective of accuracy alignment and conduct the corresponding analysis.

\subsection{Analysis from a Perspective of Accuracy Alignment}\label{FFWP}
Previously, we have identified that the clean and robust accuracy of partial classes is not aligned, which means that the clean accuracy of these classes does not translate to the corresponding level of robust accuracy. To investigate the relationship between clean and robust accuracy as a whole, we analyze from a perspective of accuracy alignment that categorizes examples into four distinct groups. This allows us to isolate and analyze instances where clean and robust accuracy does not correlate proportionally. Based on this analysis, we leverage these observations and patterns to develop FAT methods for enhancing model robustness. Note that the experiment settings are aligned with the situation from the above class-wise perspective.
\subsubsection{Definition of the Perspective of Accuracy Alignment}
\begin{figure}[t]\centering
	\includegraphics[scale=0.48]{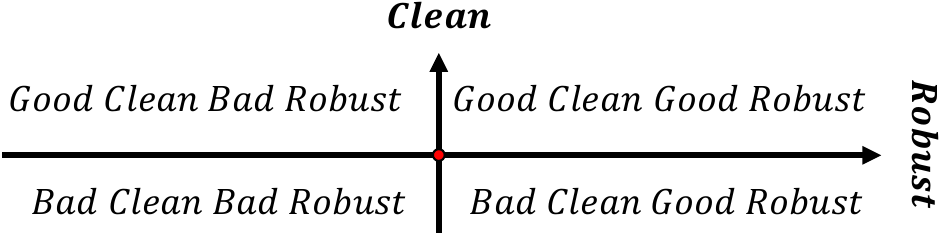}
	\caption{Schematic diagram of the perspective of accuracy alignment taxonomy.}
	\label{KWImgs}
\end{figure}

\begin{figure}[t]\centering
	\includegraphics[scale=0.26]{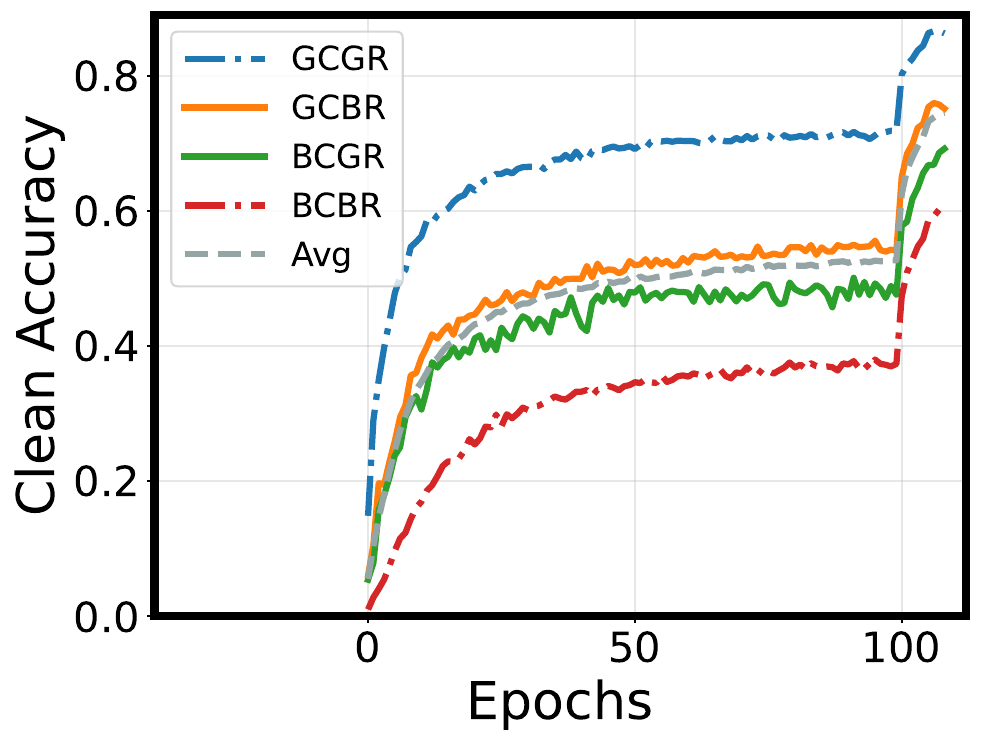}
	\includegraphics[scale=0.26]{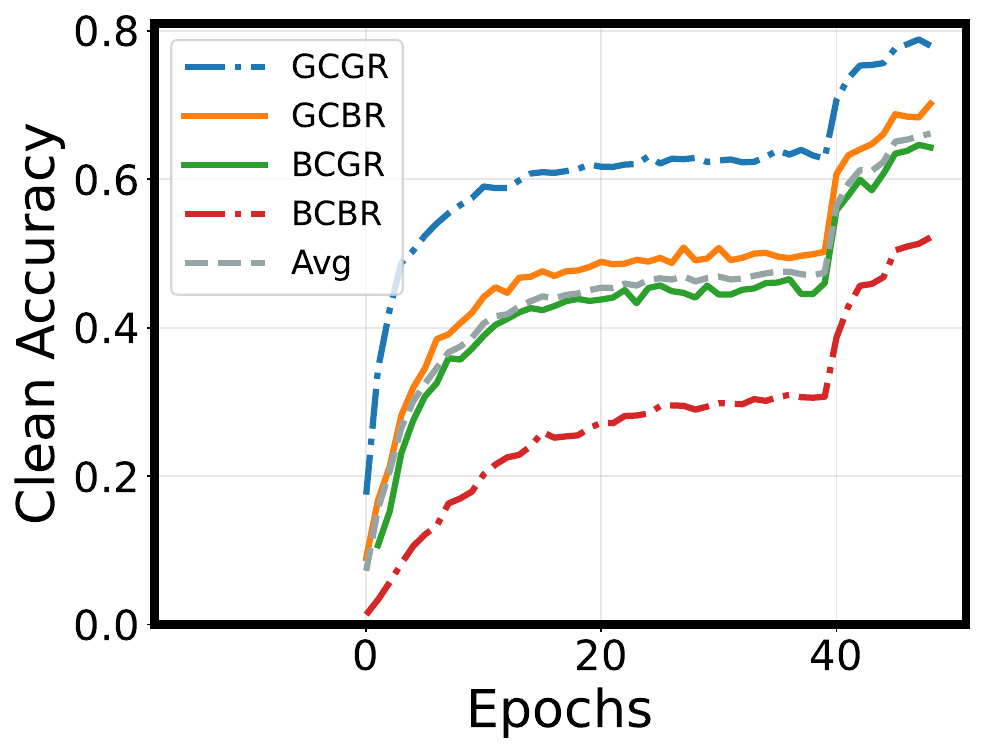}
	\includegraphics[scale=0.26]{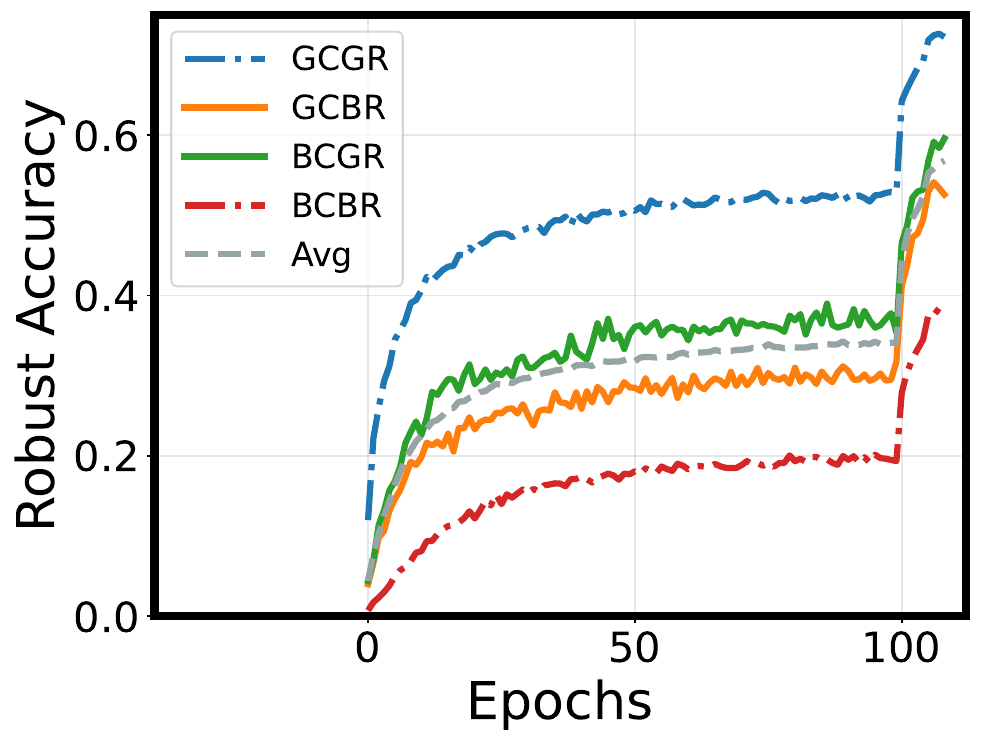}
	\includegraphics[scale=0.26]{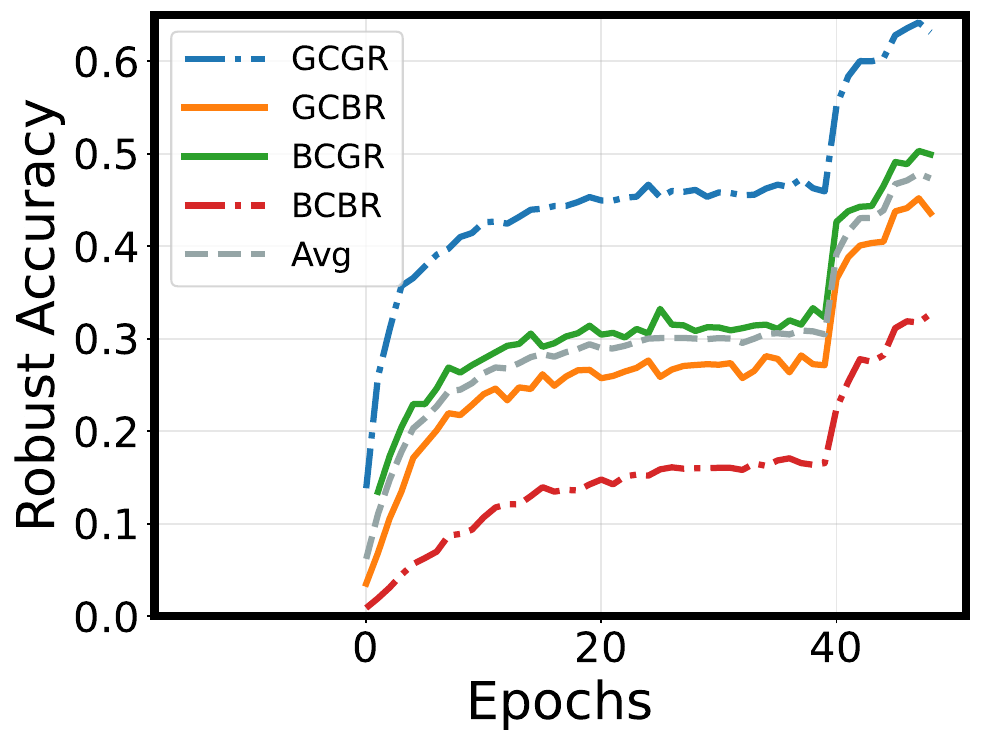}
	\subfigure[]{\includegraphics[scale=0.26]{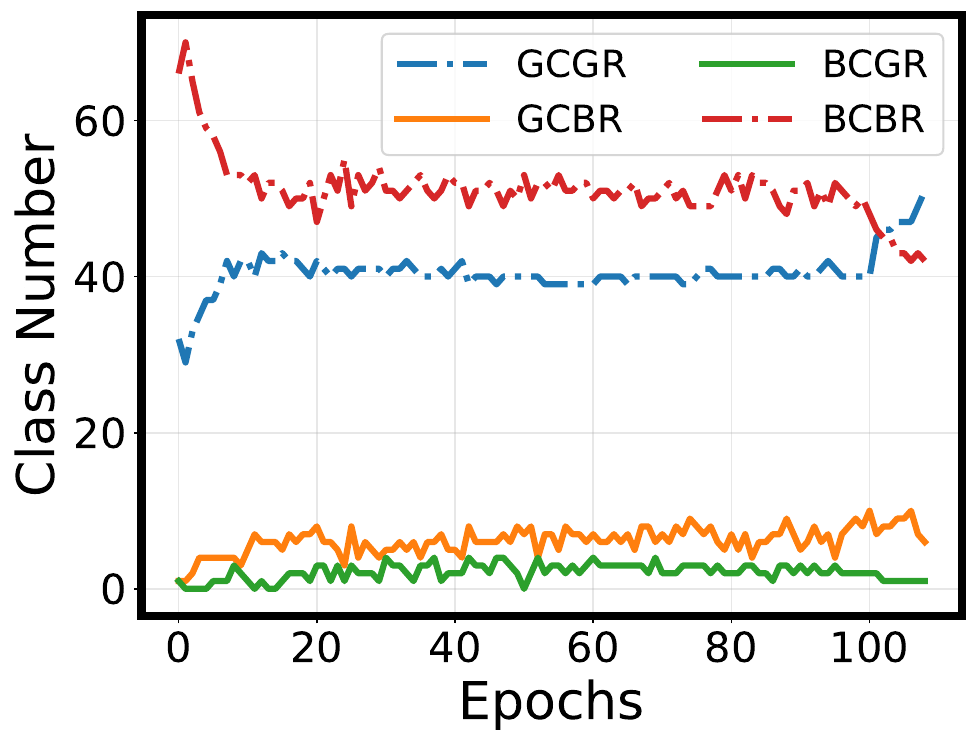}}
	\subfigure[]{\includegraphics[scale=0.26]{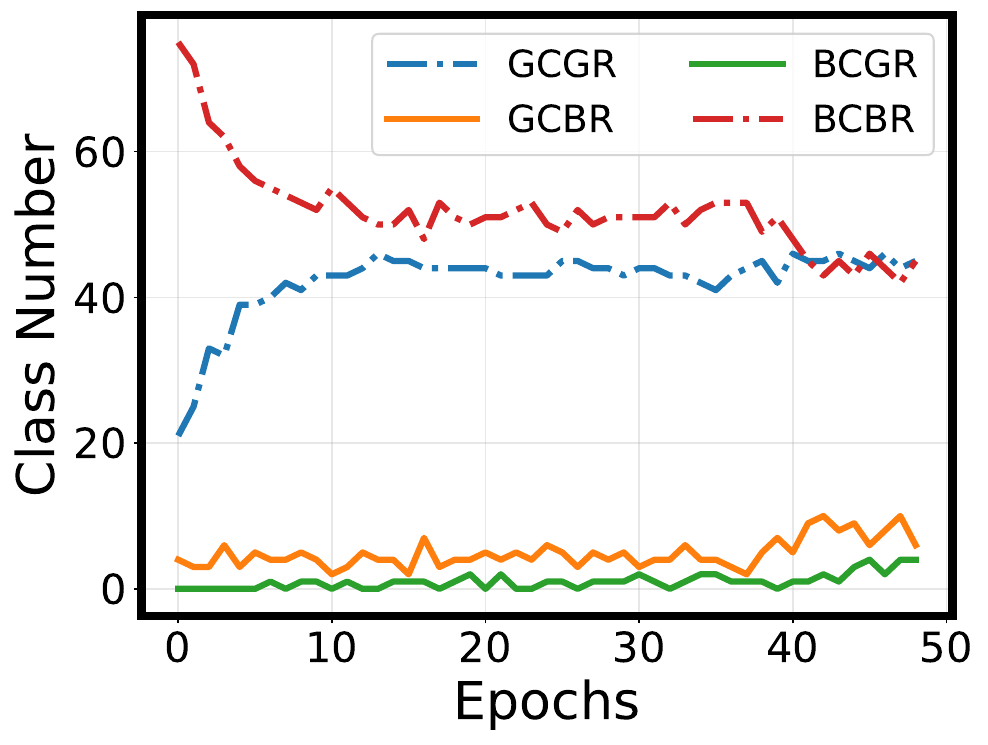}}
	\caption{Training results under the perspective of accuracy alignment using FGSM-RS on CIFAR-100 and ImageNet-100. ``Avg'' denotes average clean or robust accuracy. The bottom row shows the number of classes in different groups that change with training. (a) Results on CIFAR-100. (b) Results on ImageNet-100.}
	\label{KWAcc}
\end{figure}

The following definitions are provided to introduce the perspective of accuracy alignment. First, the perspective index $(a^i_c, a^i_r)$ is formulated as $(a^i_c, a^i_r) = \big{(}(c_i-c), (r_i-r)\big{)}$, where $a^i_c$ and $a^i_r$ denote the clean and robust classification coefficients of the $i$-th class, respectively. Meanwhile, $c_i$, $r_i$, $c$, and $r$ represent the $i$-th class clean accuracy, $i$-th class robust accuracy, overall clean accuracy, and overall robust accuracy, respectively. Accordingly, the examples are divided into four groups in Fig. \ref{KWImgs}. 
Subsequently, the good clean and good robust (GCGR) class is defined as $(a^i_c > 0, a^i_r > 0)$; the good clean and bad robust (GCBR) class for $(a^i_c > 0, a^i_r < 0)$; the bad clean and good robust (BCGR) class for $(a^i_c < 0, a^i_r > 0)$; the bad clean and bad robust (BCBR) class for $(a^i_c < 0, a^i_r < 0)$. The training information of FGSM-RS on CIFAR-100 and ImageNet-100 training sets from the perspective of accuracy alignment is presented in Fig. \ref{KWAcc}. As shown in Fig. \ref{KWAcc}, the clean and robust accuracy of some examples is not strictly positively correlated, as is observed in the cases of GCBR and BCGR. Therefore, we further investigate and leverage this result to enhance the performance of FAT. 

\subsubsection{Observations and Analysis}
First, the model exhibits favorable clean and robust accuracy for classes in GCGR, indicating that the model effectively learns the features of examples for classification. This leads to output predictions for clean examples and AEs that are more closely aligned. In contrast, the model shows low clean and robust accuracy for classes in BCBR, suggesting that the model does not learn the features of BCBR examples well, resulting in a significant difference between the predictions for clean examples and AEs. Meanwhile, both clean and robust accuracy for the GCBR and BCGR classes are close to the average accuracy, indicating that the model extracts effective classification features. Furthermore, the robust accuracy of GCBR and BCGR classes is much better than that of BCBR classes (+14\% for GCBR and +18\% for BCGR) on CIFAR-100, indicating that the model successfully learns some of the robust features of these classes during training. Therefore, the learning difficulty of the model for these classes is lower than that for hard examples in BCBR classes. Further learning the robust features of these classes and improving overall robust accuracy is promising, making it necessary to emphasize these examples during training. 

Additionally, both GCBR and BCGR classes exhibit better clean accuracy than robust accuracy, indicating that their clean accuracy provides more positive guidance compared to that of BCBR classes. Leveraging their clean accuracy as additional guidance to enhance their robust accuracy is meaningful. Based on these observations, our goal is to improve FAT performance through a regularization method that utilizes naturally generated metrics during training. We achieve this by aligning the outputs for clean examples and AEs through regularization and adjusting the regularization strength according to the differences among the four groups. The strategy for regularization strength across the four groups varies based on their classification performance. For GCGR, which demonstrates good classification accuracy during training, the regularization intensity is reduced to focus solely on maintaining the current performance of the model. In contrast, for BCBR, where both clean and robust accuracy are relatively low, aligning the predictions for clean examples and AEs leads to instability without contributing to improved training. Thus, the regularization intensity for BCBR is also should decreased. For GCBR and BCGR, their robust features have the potential to be further developed to enhance overall robustness. Therefore, regularization is intensified for these groups to better align the outputs of clean examples and AEs. This approach aims to guide improvements in robust accuracy by promoting more reliable predictions for clean examples.

\subsubsection{Class Transition Analysis}
\begin{figure}[t]\centering
	\subfigure[Transition of classes with GCBR]{\includegraphics[scale=0.3]{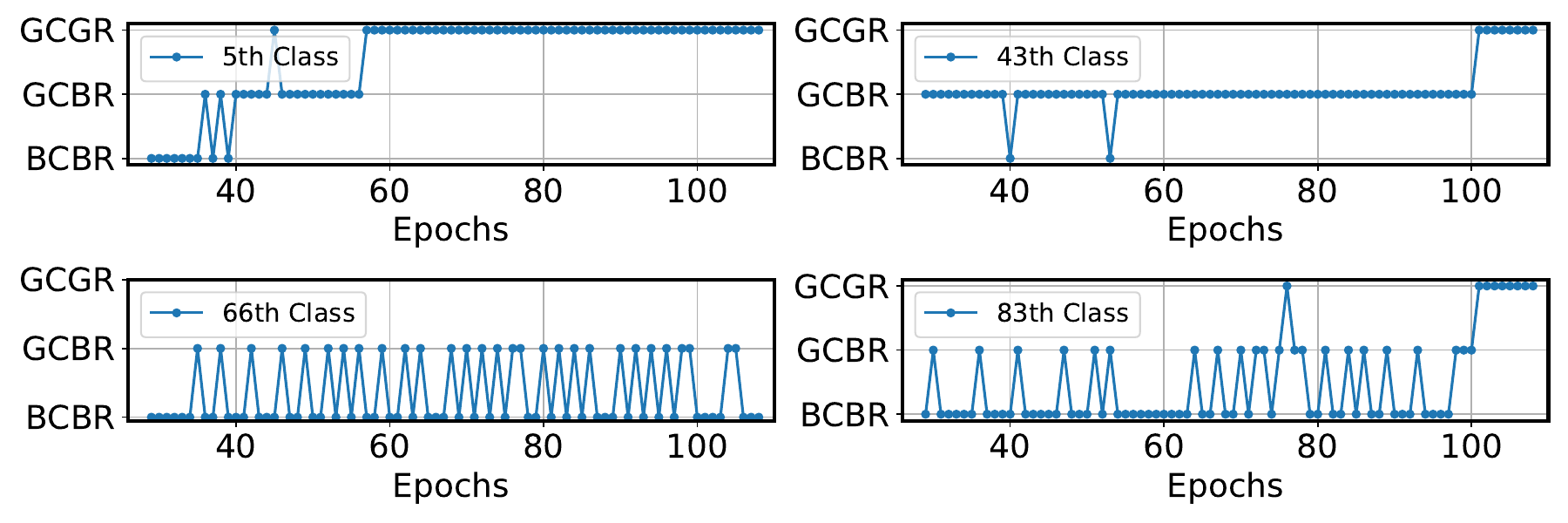}}
    \subfigure[Transition of classes with BCGR]{\includegraphics[scale=0.3]{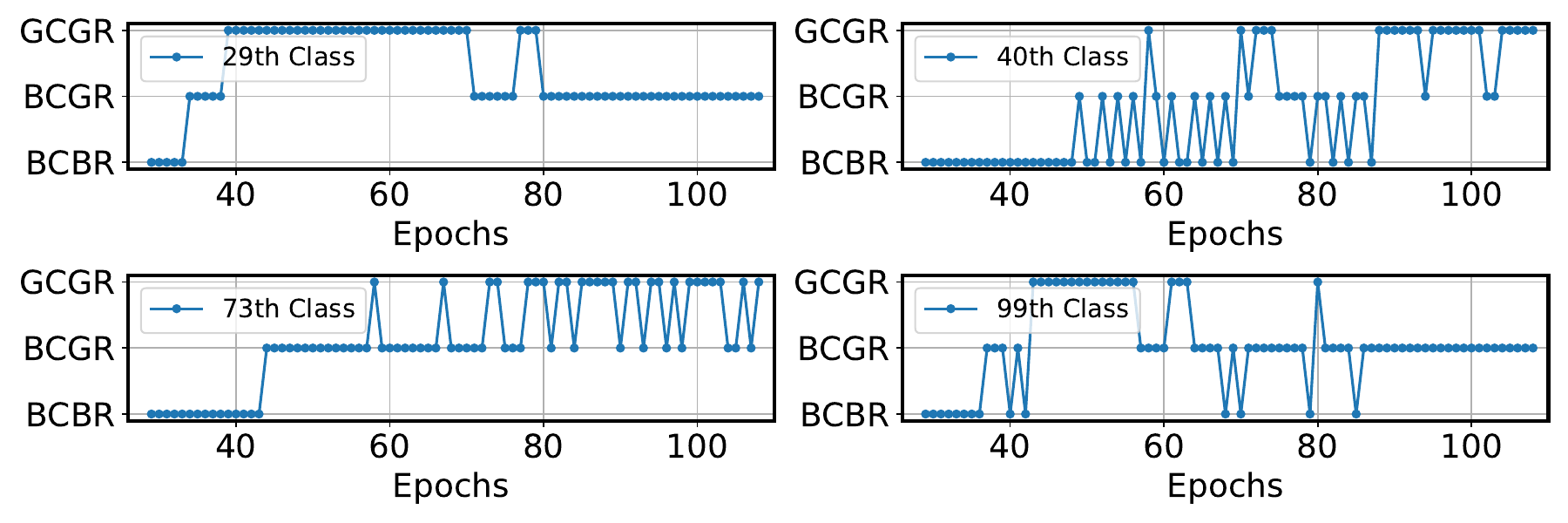}}
	\caption{Transition of classes with GCBR or BCGR on CIFAR100 dataset during FGSM-RS training.}
	\label{BCGRTrans}
\end{figure}
To analyze the transition trends of classes among groups from the perspective of accuracy alignment, we select several representative classes from the CIFAR-100 for in-depth analysis. Since classes belonging to GCGR and BCBR remain stable and rarely undergo transitions during training, our analysis focuses on the transition trends of classes in GCBR or BCGR. We begin this analysis from the 30th epoch to ensure that the training is stabilized. The corresponding results are illustrated in Fig. \ref{BCGRTrans}. For classes in GCBR, clean accuracy improves faster than robust accuracy, leading to oscillations between GCBR and BCBR. As training progresses, the robust accuracy of some classes increases, eventually reaching GCGR. Classes that frequently appear in BCGR exhibit similar oscillation patterns to those in GCBR, and are easier to improve to reach GCGR. By comparing the frequency difference between GCBR and BCGR classes being converted into GCGR classes, classes belonging to BCGR are easier and faster to reach GCGR than the classes belonging to GCBR, which indicates that clean accuracy is easier to improve. Moreover, note that when a class undergoes transition due to changes in accuracy, typically, each transition of one class only involves clean or robust accuracy. For example, classes in GCGR generally convert to GCBR or BCGR each time. In other words, the transition of classes among four groups involves a step-by-step process. These observations lead to two conclusions. First, the oscillations suggest that the model's classification accuracy for GCBR and BCGR classes hovers around the borderline, causing the accuracy of these classes to fluctuate around the average accuracy during optimization. This insight suggests potential strategies to enhance overall training performance by emphasizing the importance of these classes during training, thereby facilitating their transition to GCGR. Second, since clean or robust accuracy transitions occur progressively, and clean accuracy is easier to improve than robust accuracy, it is essential to leverage clean output to guide robust output, which can further enhance training performance.

\subsubsection{Analysis on Different Models and Large Dataset}
\begin{figure}[t]\centering
	\subfigure[DenseNet121]{\includegraphics[scale=0.26]{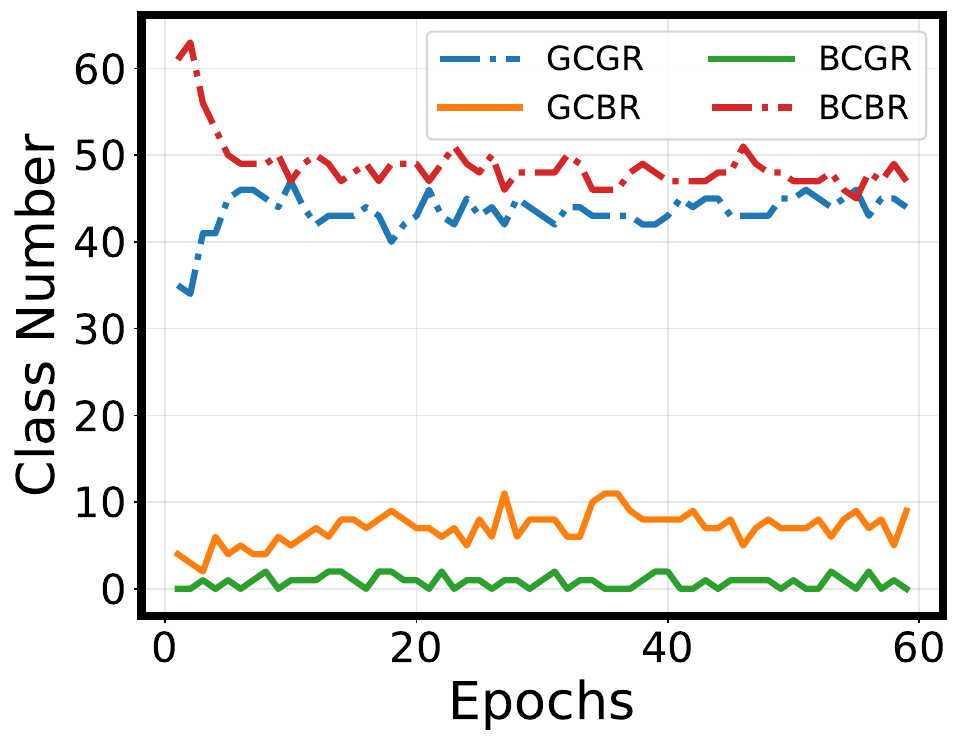}}
	\subfigure[InceptionV3]{\includegraphics[scale=0.26]{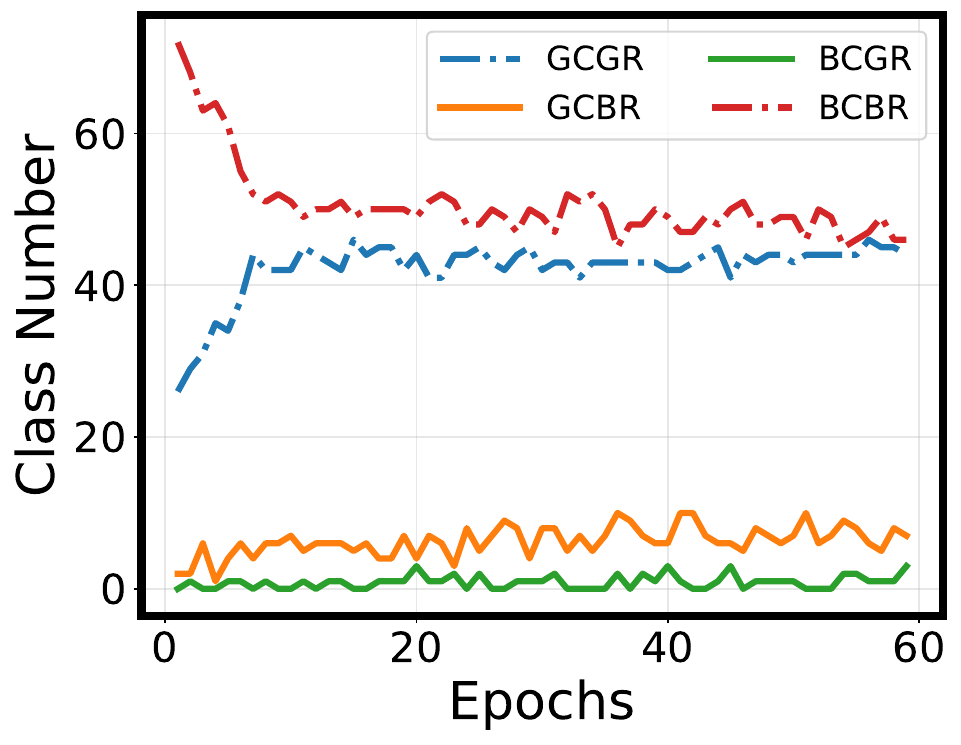}}
	\subfigure[Swin Transformer]{\includegraphics[scale=0.26]{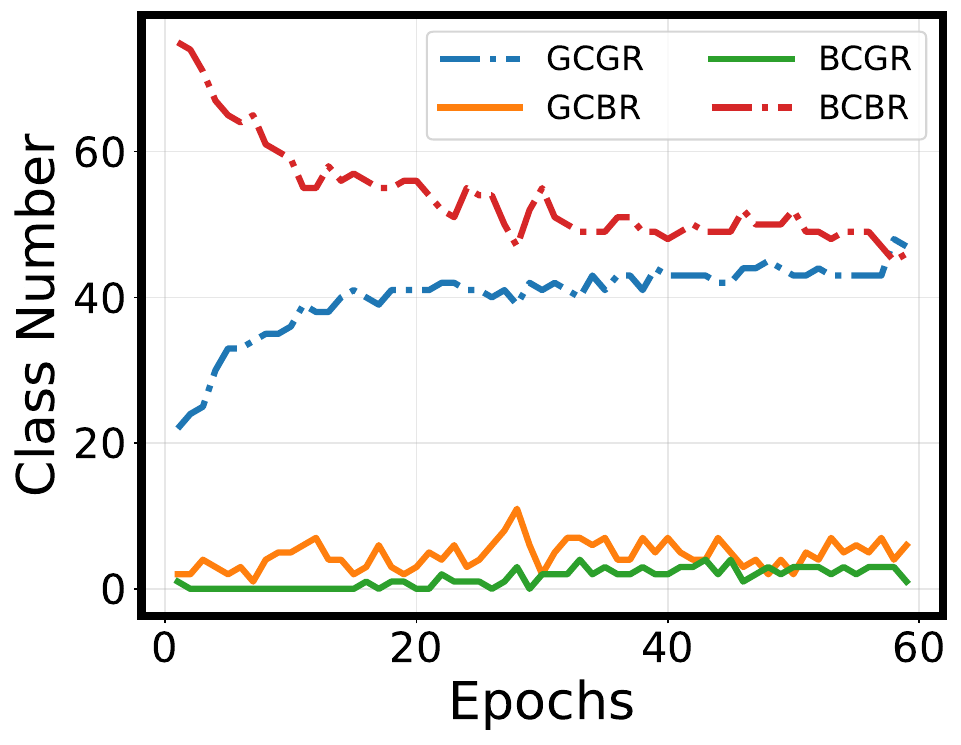}}
    \subfigure[ResNet50 on ImageNet-1K]{\includegraphics[scale=0.26]{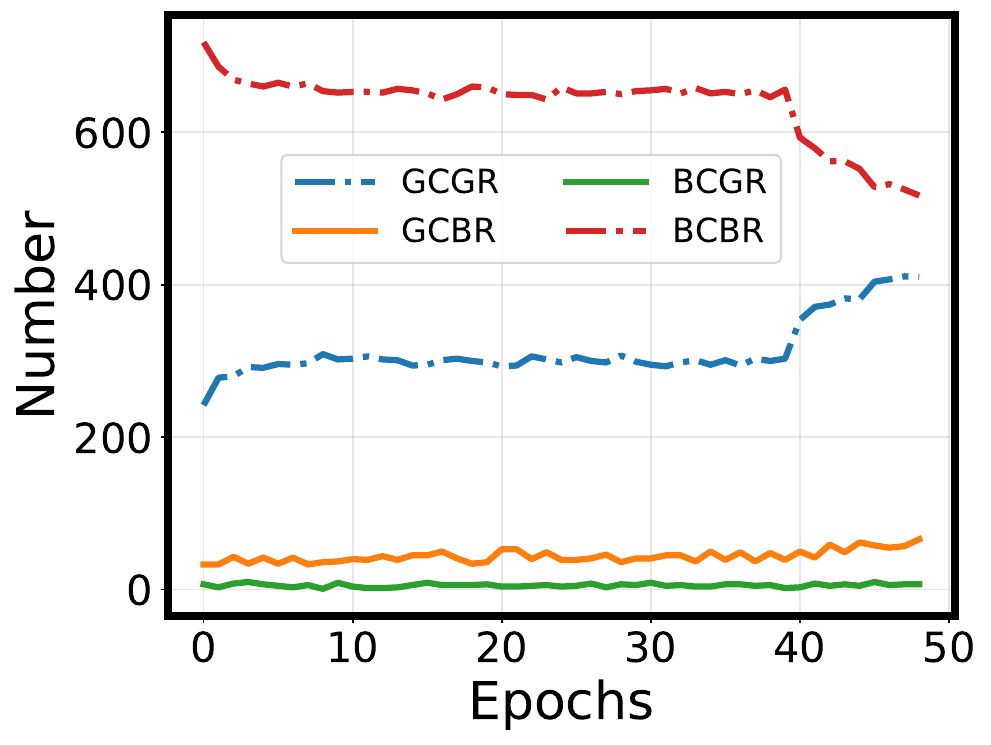}}
	\caption{The number of classes included in each group of the perspective of accuracy alignment during FGSM-RS training. (a)-(c) CIFAR-100 results with different models. (d) Results on ImageNet-1K with ResNet-50. }
	\label{DiffRSKWNum}
\end{figure}
To validate the universality of our perspective of accuracy alignment analysis, experiments are extended and performed on ImageNet-1K. Furthermore, we also evaluate various models on the CIFAR-100 dataset. The results for CIFAR-100 with different models are presented in Figs. \ref{DiffRSKWNum} (a) to (c), while the results for ImageNet-1K using ResNet50 are shown in Fig. \ref{DiffRSKWNum} (d). Specifically, we consider DenseNet-121 \cite{DenseNet}, Inception V3 \cite{IncV}, and Swin Transformer \cite{SwinTrans} for evaluation. For DenseNet121 and Inception V3, we employ training methods consistent with those outlined in the manuscript. For the Swin Transformer Tiny, we use Adam optimizer with a learning rate of $\sigma=0.0001$. The results indicate that, across different models and the large-scale dataset, the analysis reveals patterns consistent with those observed in results using ResNet-18.

\subsection{Discussion and Motivation of Methodology}
Empirical results from both class-wise and accuracy alignment perspectives reveal two key conclusions. First, FAT encounters a disparity between clean and robust accuracy across different class examples, with clean accuracy surpassing robust accuracy. Second, there are certain examples where clean and robust accuracy is not strictly positively correlated. Therefore, we aim to leverage these differences and observations to enhance robustness by aligning the model's clean and robust outputs, using clean outputs to guide robust outputs. Additionally, we employ differentiated training configurations from class-wise and accuracy alignment perspectives to improve performance. However, introducing additional hyperparameters to achieve this results in an unacceptably complex tuning process. This challenge motivates us to develop SKG-FAT, a method designed to enhance FAT by utilizing naturally occurring indicators during training without the requirements for additional hyperparameters.

\section{Methodology}\label{Met}
\begin{algorithm}[t]
	\renewcommand{\algorithmicrequire}{\textbf{Input:}}
	\renewcommand{\algorithmicensure}{\textbf{Output:}}
	\caption{Implementation of SKG-FAT with CWR}
	\label{algOne}
	\begin{algorithmic}[1]
		\State \textbf{Input}: Neural network $f_\theta$ with parameter $\theta$; Clean examples $\vec{x}$ and label $\vec{y}$; Learning rate $\sigma$; Regularization hyperparameter $\lambda$; Minimum relaxation factor $\kappa_{\text{min}}$; Perturbation initialization $\vec{\delta}_0$; Perturbation budget $\epsilon$.
		\State \textbf{Return}: A robust neural network $\bar f_\theta$.
		\State Initialization: $\theta \gets \theta_0$ and $\vec{y}^i_k \gets \vec{y}$.
		\For{$k$ in $EPOCHs$}
		\State $c^i_k \gets \text{ClassWiseClean}(f_{\theta}(\vec{x}))$.
			\For{$batch$ in $BATCHs$}
				\State $\vec{\delta} \gets \Phi_{\epsilon}\Big{(}\epsilon\cdot\text{sign}(\nabla_{\vec{x}+\vec{\delta}_0}\mathcal{L}\big{(}f_{\theta}(\vec{x}+\vec{\delta}_0), \vec{y}^i_k\big{)}\Big{)}$.
				\State $\mathcal{L}_{\text{adv}} \gets \mathcal{L}\big{(}f_{\theta}(\vec{x}+\vec{\delta}), \vec{y}^i_k\big{)}$.
				\For{$i$ in $m$}						
                    \State $\Omega_{\text{CWR}} \gets$ equation \eqref{CWRFormula}.
				\EndFor
				\State $\mathcal{L}_{\text{OA}} \gets \mathcal{L}_{\text{adv}} + \lambda \cdot \Omega_{\text{CWR}}$.
				\State $\theta \gets \theta - \sigma\nabla_{\theta}\mathcal{L}_{\text{OA}}$.
			\EndFor
		\State $\vec{y}^i_k\gets \kappa(c^i_k)\cdot\vec{y}+(\vec{y}-1)\cdot \frac{\kappa(c^i_k)-1}{m-1}$.
		\EndFor
	\end{algorithmic}  
\end{algorithm}


\subsection{Self-Knowledge Guided Regularization}
Using regularization to improve FAT has proven to be an effective approach \cite{pang2022robustness}. First, the mathematical representation of objective minimax optimization for adversarial training with regularization $\Omega$ is introduced and formulated as follows:
\begin{equation}
	\min_{\theta} \mathbb{E}_{\vec{x}\sim\mathcal{D}}\Big{[}\Omega+\max_{\vec{\delta}\in\Lambda} \mathcal{L}\Big{(}f_{\theta}(\vec{x}+\vec{\delta}), \vec{y}\Big{)}\Big{]}.
\end{equation}
In this paper, we develop our method based on MSE regularization for two primary reasons. On the one hand, MSE requires only the inference results for clean examples and AEs, which can ensure training efficiency \cite{TDAT}. In contrast, regularization in GradAlign requires computing the gradients with respect to clean examples and AEs, resulting in double training time \cite{GradAlign, NFGSM}. On the other hand, Pang $et~al.$ demonstrate that substituting KL divergence with distance metrics can achieve a better trade-off between robustness and clean accuracy \cite{pang2022robustness}. Then, the formula for MSE regularization is recalled as follows:
\begin{equation}\label{MSEDef}
	\Omega = \lambda \cdot \frac{1}{n}\sum^n_{u=1}\big{(}f_{\theta}(\vec{x}^u) - f_{\theta}(\vec{x}^u+\vec{\delta})\big{)}^2, 
\end{equation}
where the hyperparameter $\lambda$ controls the penalty strength of regularization. This method minimizes the distance between the probability vectors of clean examples and AEs, providing a more reliable and stable update direction \cite{GAT}. Consequently, it results in stronger attacks that enhance training effectiveness. As observed in subsection \ref{FCWP}, datasets exhibit significant accuracy differences during the training process. Rather than being overlooked, these differences should be effectively harnessed to improve model robustness. Motivated by these technology gaps, we develop new methods to achieve two key objectives for improving training performance. First, given that clean accuracy is higher than robust accuracy, we aim to enhance robust accuracy by minimizing the outputs for clean examples and AEs. Second, considering the performance differences across the class-wise and accuracy alignment perspectives, we determine the regularization strength based on these differences without introducing extra hyperparameters. To accomplish these objectives, we utilize natural metrics derived from the training to develop new regularization.

\subsubsection{Class-Wise Guided Regularization}\label{CWER}
Motivated by the difference in class-wise perspective as presented in subsection \ref{FCWP}, we introduce class-wise guided regularization (CWR) $\Omega_{\text{CWR}}$, which is formulated as follows:
\begin{equation}\label{CWRFormula}
	\Omega_{\text{CWR}} = \lambda \cdot \frac{1}{n} \sum^m_{i=1}c_i\sum^{n/m}_{u=1}\big{(}f_{\theta}(\vec{x}^u_i) - f_{\theta}(\vec{x}^u_i+\vec{\delta})\big{)}^2, 
\end{equation}
where $\vec{x}_i$ represents the example from the $i$-th class, and $c_i$ denotes the CWR guidance factor of the $i$-th class in the training set. We aim to implement the CWR guidance factor that can be applied to different datasets and models without introducing additional parameters. To accomplish this, a simple yet effective approach based on the clean accuracy of the corresponding class is employed, driven by three key considerations. First, clean accuracy generally exceeds robust accuracy and is easier to improve, making it a more reliable measure for differentiation. Second, when training with datasets containing numerous classes, the robust accuracy of partial classes may be zero or close to zero. This makes $c_i$ ineffective for guiding regularization strength, whereas using clean accuracy avoids this issue. That is to say, clean accuracy can better reflect the differences in regularization strength between different classes. Furthermore, classification accuracy naturally possesses upper and lower bounds, it eliminates the need for hyperparameter adjustments. Therefore, the overall loss function $\mathcal{L}_{\text{OA}}$ with CWR for performing training is presented as $\mathcal{L}_{\text{OA}}=\mathcal{L}_{\text{CE}}+\Omega_{\text{CWR}}$, where $\mathcal{L}_{\text{CE}}$ signifies the standard cross entropy loss. The integration of CWR enhances the performance of FAT by more effectively managing the impact of regularization across diverse classes. The pseudocode of our SKG-FAT with CWR is presented in Algorithm \ref{algOne}.
\subsubsection{Why CWR Can Improve Robustness}
The proposed CWR aims to minimize the discrepancy between the clean prediction for each class and its corresponding prediction for AEs according to class-wise training state. This approach leverages the mechanism that classes with higher clean accuracy indicate that the model has learned reliable classification features \cite{AEE}. Meanwhile, since clean accuracy exceeds robust accuracy, CWR enhances the alignment between clean outputs and AE outputs for these high-accuracy classes, thereby utilizing the clean outputs to guide the model output of AEs and improve robustness. Conversely, for classes with lower accuracy, the model has not yet effectively learned the classification features. In such cases, enforcing regularization to align clean outputs with AE outputs is less meaningful, as the guidance from low clean and robust accuracies is ineffective. Therefore, the regularization strength is reduced for these lower accuracy classes. By furnishing CWR, we assign differentiated training strategies to different classes, ultimately improving the overall robust accuracy of the model.

\begin{algorithm}[t]
	\renewcommand{\algorithmicrequire}{\textbf{Input:}}
	\renewcommand{\algorithmicensure}{\textbf{Output:}}
	\caption{Implementation of SKG-FAT with AGR}
	\label{algTwo}
	\begin{algorithmic}[1]
		\State \textbf{Input}: Neural network $f_\theta$ with parameter $\theta$; Clean examples $\vec{x}$ and label $\vec{y}$; Learning rate $\sigma$; Regularization hyperparameter $\lambda$; Minimum relaxation factor $\kappa_{\text{min}}$; Perturbation initialization $\vec{\delta}_0$; Perturbation budget $\epsilon$.
		\State \textbf{Return}: A robust neural network $\bar f_\theta$.
		\State \textbf{Initialization}: $\theta \gets \theta_0$, $\vec{y}^i_k \gets \vec{y}$, $\vec{\delta}_0 \gets \mathcal{U}(-\epsilon, \epsilon)$, good clean list $L_{\text{GC}}$, bad clean list $L_{\text{BC}}$, good robust list $L_{\text{GR}}$, and bad robust list $L_{\text{BR}}$.
		\For{$k$ in $EPOCHs$}
		\State $(a^i_c, a^i_r) \gets \big{(}(c^i_{k-1}-c_{k-1}), (r^i_{k-1}-r_{k-1})\big{)}$,
		\For{$i$ in $m$}
			\IfThenElse {$a^i_c > 0$} {$L_{\text{GC}} \gets i$} {$L_{\text{BC}} \gets i$}
			\IfThenElse {$a^i_r > 0$} {$L_{\text{GR}} \gets i$} {$L_{\text{BR}} \gets i$}
		\EndFor
		\State GCGR $\gets L_{\text{GC}} \cap L_{\text{GR}}$, GCBR $\gets L_{\text{GC}} \cap L_{\text{BR}}$
		\State BCGR $\gets L_{\text{BC}} \cap L_{\text{GR}}$, BCBR $\gets L_{\text{BC}} \cap L_{\text{BR}}$
\For{$batch$ in $BATCHs$}
		\State $\vec{\delta} \gets \text{Clip}_{\epsilon}\Big{(}\epsilon\cdot\text{sign}(\nabla_{\vec{x}+\vec{\delta}_0}\mathcal{L}\big{(}f_{\theta}(\vec{x}+\vec{\delta}_0), \vec{y}^i_k\big{)}\Big{)}$.
		\State $\mathcal{L}_{\text{adv}} \gets \mathcal{L}\big{(}f_{\theta}(\vec{x}+\vec{\delta}), \vec{y}^i_k\big{)}$.
		\State $L_{\text{t}} \gets [$GCGR, GCBR, BCGR, BCBR$]$.
		\For{$L^{j}_{\text{t}}$ is in $L_{\text{t}}$}
			\State $f^{j}_{\theta}(\vec{x}) \gets \text{IndexSelect}(f_{\theta}(\vec{x}), L^{j}_{\text{t}})$
			\State $f^{j}_{\theta}(\vec{x}+\vec{\delta}) \gets \text{IndexSelect}(f_{\theta}(\vec{x}+\vec{\delta}), L^{j}_{\text{t}})$
			\State $\Omega_{\text{AGR}} \gets $ equation \eqref{AGRFormula}.
		\EndFor
		\State $\mathcal{L}_{\text{OA}} \gets \mathcal{L}_{\text{adv}} + \lambda \cdot \Omega_{\text{AGR}}$
		\State $\theta \gets \theta - \sigma\nabla_{\theta}\mathcal{L}_{\text{OA}}$
	\EndFor
	\State $c_k \gets \text{AvgClean}(f_{\theta}(\vec{x}))$
	\State $c^i_k \gets \text{ClassWiseClean}(f_{\theta}(\vec{x}))$
	\State $r_k \gets \text{AvgRobust}(f_{\theta}(\vec{x}+\vec{\delta}))$
	\State $r^i_k \gets\text{ClassWiseRobust}(f_{\theta}(\vec{x}+\vec{\delta}))$
	\State $\vec{y}^i_k \gets \kappa(c^i_k)\cdot\vec{y}+(\vec{y}-1)\cdot \frac{\kappa(c^i_k)-1}{m-1}$.
\EndFor
	\end{algorithmic}  
\end{algorithm}

\subsubsection{Accuracy Alignment Guided Regularization}
Then, we leverage the analysis results from the perspective of accuracy alignment to introduce the accuracy alignment guided regularization (AGR), which effectively mitigates the adverse effects of differences in groups from the perspective of accuracy alignment. Considering the classes are divided into four groups and displaying distinct accuracy situations. Consequently, the mathematical description of the AGR $\Omega_{\text{AGR}}$ is formulated as
\begin{equation}\label{AGRFormula}
	\Omega_{\text{AGR}} = \lambda \cdot \frac{1}{n} \sum^4_{j=1}d_j\sum^{n/4}_{u=1}\big{(}f_{\theta}(\vec{x}^u_j) - f_{\theta}(\vec{x}^u_j+\vec{\delta})\big{)}^2, 
\end{equation}
where $j$ denotes the $j$-th groups in the perspective of accuracy alignment as defined in subsection \ref{FFWP} and $d_j$ represents the AGR guidance factor. From the perspective of accuracy alignment, the number of categories is reduced compared to the situation in the class-wise perspective, such as reducing 100 classes in the CIFAR-100 to 4 groups. Meanwhile, according to the analysis in subsection \ref{FFWP}, due to the similar clean accuracy between the GCBR group and BCGR group, and the significant reduction in the number of regularization terms caused by the four groups, choosing the clean accuracy of each group as the regularization guidance factor could not provide enough regularization strength difference. Consequently, the differentiation in regularization strength for each group would not be significant enough, limiting performance improvement. Nonetheless, the differentiation in the number of groups in the perspective of accuracy alignment exceeds the accuracy differences. As shown in Fig.\ref{KWAcc}, the differences in the number of classes in each group are significant, this provides an opportunity for the reasonable implementation of AGR. Therefore, we use the number of classes in each group of the perspective of accuracy alignment to implement the AGR guidance factor:

\begin{equation}\label{AGRImpl}
	d_j = 1/n_j,
\end{equation}
where $n_j$ signifies the number of $j$-th groups. Similarly, the overall loss $\mathcal{L}_{\text{OA}}$ with AGR is presented as $\mathcal{L}_{\text{OA}}=\mathcal{L}_{\text{CE}}+\Omega_{\text{AGR}}$. This regularization can enhance performance almost without introducing additional training time consumption, even if the number of classes in the dataset is large. The pseudocode for our SKG-FAT with CWR is presented in Algorithm \ref{algTwo}.

\subsubsection{Why AGR Can Improve Robustness}
AGR adjusts the impact of examples from different classes on training by considering the number of classes in each group, whereas the standard MSE loss neglects this variability. Specifically, the definition of MSE is introduced in equation \eqref{MSEDef}, where $1/n$ can be seen as a uniform parameter for all examples. In contrast, AGR adapts group weights according to the number of classes within each group. As in equation \eqref{AGRImpl}, $1/n_j$ assigns variable weights across groups based solely on training-derived information, with $\lambda$ as the only hyperparameter, which is typical in regularization methods. Furthermore, subsection \ref{FFWP} suggests that regularization for GCGR and BCBR should be less emphasized while enhancing the weights for GCBR and BCGR. This adjustment is intuitive because GCGR and BCBR constitute most of the examples, whereas GCBR and BCGR are less frequent  (as Fig. \ref{KWAcc}(bottom) and Fig. \ref{DiffRSKWNum}). Thus, AGR dynamically adjusts the weights for examples from the four groups, effectively configuring their influence in training.

\subsection{Self-Knowledge Guided Label Relaxation}\label{LERDef}
It has been identified that catastrophic overfitting in FAT is linked to the imbalance of inner and outer optimization in the min-max problem \eqref{ATDefine} \cite{TDAT}. Existing methods demonstrate that stabilizing classification confidence \cite{TDAT} or preventing the over-memorization \cite{linover} can mitigate catastrophic overfitting and improve robust accuracy. Motivated by our findings of class-wise differences in example accuracy, we propose self-knowledge guided label relaxation (SKLR) as
\begin{equation}\label{LER}
	\vec{y}^i_k=\kappa^i_k\cdot\vec{y}+(\vec{y}-1)\cdot \frac{\kappa^i_k-1}{m-1},
\end{equation}
where $\kappa^i_k$ denotes the class-wise label relaxation factor, which changes with the training performance. Consistent with the reasons involved in our CWR, clean accuracy is easier to improve and eliminates the requirements for hyperparameter upper and lower bound adjustments. Therefore, we exploit the knowledge of clean accuracy to implement the class-wise label relaxation factor to guide training as
\begin{equation}
	\kappa^i_k=\min\big{(}\max(|\log c^i_k|, \kappa_{\min}), 1\big{)},
\end{equation}
where $c^i_k$ denotes the clean accuracy of the $i$-th class on training set in $k$-th training epoch, $\kappa_{\min}\in((1/m), 1]$ is minimum relaxation factor, $\vec{y}$ denotes the one-hot label, and $\vec{y}^i_k$ represents the relaxation label of $i$-th class in the $k$ training epoch. SKLR dynamically adjusts labels based on the training state in a class-wise manner, effectively mitigating inner-outer optimization imbalance. This improves training performance and alleviates catastrophic overfitting. 
\par
The imbalance optimization of the minimax in FAT causes catastrophic overfitting. SKLR addresses this by adjusting the strength of label relaxation based on the training status of each class, thereby stabilizing the training process. In the initial training, this method adopts higher confidence to support the model in learning the primary feature distribution of examples, thereby rapidly optimizing the accuracy. After that, this method facilitates label smoothness, encouraging the model to classify examples correctly without pursuing excessively high confidence, which contributes to preventing over-memory and thus preventing catastrophic overfitting. Overall, this approach stabilizes the minimax optimization in FAT, effectively preventing catastrophic overfitting.

\begin{table*}[t]
	\centering
	\caption{Accuracy and Training Time Comparisons and Training Time on CIFAR-10 and CIFAR-100. Bold Number Denotes Better Than Existing Methods. Underline Represents the Best Robust Accuracy.}
	\setlength{\tabcolsep}{1.1mm}{
	\begin{tabular}[l]{@{}l| c c c c c l| c c c c c l}
	\toprule[2pt]
	\multirow{3}*{Method} &\multicolumn{6}{c|}{CIFAR-10} &\multicolumn{6}{c}{CIFAR-100} \\
    \cmidrule(r){2-7} \cmidrule(r){8-13}
	&Clean &MI &PGD50 &CW &AA &\multirow{2}*{Time} &Clean &MI &PGD50 &CW &AA &\multirow{2}*{Time}\\
    &Best / Last &Best / Last &Best / Last &Best / Last &Best / Last & &Best / Last &Best / Last &Best / Last &Best / Last  &Best / Last\\
	\toprule[1pt]
    \multirow{1}*{FGSM-RS}   	&83.6 / 83.6 &48.9 / 48.9 &45.9 / 45.9 &46.0 / 46.1 &42.8 / 42.8 &51 &51.6 / 51.6 &23.0 / 23.0 &21.7 / 21.7 &20.9 / 20.9 &18.7 / 18.7 &64\\
	\multirow{1}*{GAT}       	&81.2 / 81.8 &54.4 / 54.8 &53.1 / 51.8 &49.7 / 49.3 &47.4 / 47.2 &114 &57.5 / 57.5 &29.5 / 29.6 &25.3 / 24.8 &25.3 / 24.8 &23.4 / 22.8 &119\\
	\multirow{1}*{FGSM-SDI}  	&83.5 / 83.7 &52.7 / 52.7 &50.3 / 50.0 &49.0 / 49.4 &46.3 / 46.3 &97 &58.6 / 58.6 &29.1 / 29.1 &27.7 / 27.7 &25.8 / 25.5 &23.2 / 23.1 &105\\
	\multirow{1}*{GradAlign} 	&80.4 / 80.4 &50.0 / 50.0 &47.6 / 47.6 &46.9 / 46.9 &43.9 / 43.9 &170 &54.8 / 55.2 &27.5 / 27.4 &26.2 / 26.2 &25.0 / 24.9 &22.3 / 22.1 &173\\
	\multirow{1}*{N-FGSM}    	&80.3 / 80.3 &50.7 / 50.8 &48.4 / 48.5 &47.3 / 47.3 &44.5 / 44.5 &51 &54.4 / 54.4 &27.5 / 27.5 &26.3 / 26.3  &25.1 / 25.1 &22.7 / 22.7 &66\\
    \multirow{1}*{PGI-BP}    	&83.1 / 83.0 &54.7 / 54.4 &53.2 / 53.2 &50.1 / 50.1 &47.4 / 47.1 &73 &57.5 / 57.7 &31.0 / 30.8 &28.9 / 28.8 &26.4 / 26.4 &23.6 / 23.3 &83\\
    \multirow{1}*{PGI-MEP}   	&81.7 / 81.7 &55.2 / 55.2 &54.1 / 54.0 &50.2 / 50.2 &48.2 / 48.0 &76 &58.7 / 58.8 &31.4 / 31.2 &31.0 / 30.7 &27.2 / 27.0 &25.1 / 25.0 &84\\
	\multirow{1}*{FGSM-SC}   	&82.4 / 82.3 &54.9 / 55.2 &53.8 / 53.8 &50.0 / 50.0 &47.9 / 47.8 &80 &58.9 / 59.4 &30.7 / 30.7 &29.1 / 28.8 &25.8 / 25.6 &23.7 / 23.5 &86\\
	\multirow{1}*{AEE}   	    &80.8 / 81.1 &51.2 / 50.9 &48.0 / 47.9 &47.1 / 46.9 &44.5 / 43.9 &176 &59.2 / 58.5 &23.2 / 23.0 &21.2 / 20.7 &21.8 / 21.0 &19.1 / 18.3 &190\\
	
	\rowcolor{black!10}\multirow{1}*{Our CWR} &80.6 / 80.8 &\textbf{\underline{56.9} / \underline{56.9}} &\textbf{\underline{55.7} / \underline{55.7}} &\textbf{\underline{50.8} / \underline{50.8}} &\textbf{\underline{48.6} / \underline{48.6}} &75 &57.8 / 58.0 &\textbf{32.4 / 32.4} &\textbf{32.0 / \underline{32.0}} &\textbf{\underline{27.8} / \underline{27.7}} &\textbf{\underline{25.4} / \underline{25.3}} &82\\
	\rowcolor{black!10}\multirow{1}*{Our AGR} &82.5 / 82.6 &\textbf{56.1 / 56.1} &\textbf{54.9 / 54.9} &\textbf{50.6 / 50.5} &48.1 / 48.0 &76 &57.5 / 57.9 &\textbf{\underline{32.8} / \underline{32.6}} &\textbf{\underline{32.2} / 31.9} &\textbf{27.6 / 27.4} &\textbf{25.5 / 25.2} &84\\
	\toprule[1pt]
	\multicolumn{13}{c}{\qquad \qquad \qquad Multi-step Adversarial Training}\\
	\toprule[1pt]
	\multirow{1}*{LAS-AWP} &82.9 / 82.9 &57.0 / 57.0 &55.2 / 55.2 &51.5 / 51.5 &49.4 / 49.4 &392 &58.7 / 58.7 &33.2 / 33.2 &32.2 / 32.2 &29.6 / 29.5 &27.3 / 27.3 &401\\
	\multirow{1}*{MART}    &82.0 / 82.3 &55.6 / 55.2 &53.5 / 52.6 &49.6 / 49.6 &47.7 / 47.7 &370 &55.1 / 55.1 &32.9 / 32.9 &31.5 / 31.2 &28.0 / 27.8 &26.1 / 25.7 &378\\
	\toprule[2pt]
\end{tabular}
}
\label{CIFARResults}
\end{table*}


\subsection{Comparison with AEE-AT}
Here, we outline the differences between our SKG-FAT and Advancing Example Exploitation (AEE) \cite{AEE}. The AEE divides examples into accuracy-crucial (A-C) and robustness-crucial (R-C). On this basis, the performance of existing adversarial training methods is investigated. While our SKG-FAT also uses decision space information, it differs by exploring example characteristics at the feature level through specific indicators to guide adversarial training. The differences are summarized in three aspects. 1) For problem analysis, AEE utilizes robust accuracy and an extra hyperparameter, dividing all examples into A-C and R-C. AEE proposes that if the indicator $c_i$ approaches zero, the example can be considered A-C. Meanwhile, we focus more on using naturally generated indicators during training to analyze and improve FAT. Furthermore, our research indicates that the assumption in AEE is not always valid (GCBR and BCBR). 2) For methodology, AEE enhances robustness for classes with easy-to-learn features (R-C) while emphasizing accuracy learning for classes with difficult-to-learn features (A-C). Our work minimizes interference among classes with different features as much as possible to improve performance. 3) For task objectives, AEE aims to address common issues in the entire adversarial training, such as the accuracy-robustness trade-off and overfitting. In contrast, our work utilizes relevant information to maximize robust accuracy while maintaining competitive clean accuracy.

\section{Experiments and Analysis}\label{EASec}
\subsection{Experiment Settings}
\subsubsection{Datasets and Training Details}
We evaluate and compare our method on CIFAR-10/100 \cite{cifar}, Tiny ImageNet, and ImageNet-100 \cite{imagenet}. Both our and existing methods are evaluated on ResNet18 \cite{ResNet} as default. We employ the SGD optimizer with a momentum of 0.9 and weight decay of 5e-4 for all datasets. For CIFAR-10, CIFAR-100, and Tiny ImageNet, the learning rate is initialed as 0.1 and divided by 10 at the 100-th and 105-th epoch with a total of 110 training epochs, respectively. For ImageNet-100, the learning rate is initialized as 0.1 and divided by 10 at the 40-th and 45-th epochs with a total of 50 epochs. We use perturbations from the previous batch to initialize the next batch examples \cite{PGIFGSM}. All experiments are conducted on a single NVIDIA 3090 GPU.

\subsubsection{Baselines for Compraison}
For a comprehensive evaluation, we compare our SKG-FAT with both classic and state-of-the-art FAT methods, including FGSM-RS \cite{FGSMRS}, GAT \cite{GAT}, NuAT \cite{TEEAT} FGSM-SDI \cite{SDI}, GradAlign \cite{GradAlign}, N-FGSM \cite{NFGSM}, FGSM-PGI (PGI-BP and PGI-MEP) \cite{PGIFGSM}, FGSM-SC with PGI-MEP \cite{FATSC}, and AEE with GradAlign \cite{AEE}. Additionally, to further demonstrate the competitiveness of our approach, we include comparisons with multi-step adversarial training, including LAS-AWP \cite{LASAT}, and MART \cite{MART}. 

\subsubsection{Adversarial Attacks for Evaluation}
We compare our SKG-FAT with other methods across different adversarial attacks, covering MIFGSM (MI) \cite{MIFGSM}, PGD-10/50 \cite{PGD}, C$\&$W (CW) \cite{CW}, and AutoAttack (AA) \cite{AutoAttack}. AutoAttack is evaluated using its official implementation. For MIFGSM, PGD, and CW attacks, we utilize the TorchAttacks \cite{torchattacks} with default settings for implementation. Moreover, the performance at both the best (selected by PGD-10) and the last checkpoint is reported to provide a comprehensive comparison \cite{PGK, CFA}.


\begin{table}[t]
    \centering
    \caption{Performance Comparisons and Training Time using ResNet18 on the CIFAR-10. All Methods Adopt the Cyclic Learning Rate for Training. Bold Numbers Denote Better Than Previous Methods. Underline Represents the Best Robust Accuracy.}
    \begin{tabular}{l|cccc}
    \toprule[1.5pt]
    \multirow{2}*{Method} & Clean & PGD-50 & AA & \multirow{2}*{Time (min)} \\
    &Best / Last &Best / Last &Best / Last \\
    \midrule
    FGSM-RS   &83.7 / 83.7 &46.1 / 46.1 &42.9 / 42.9 & 15 \\
    FGSM-SDI  &82.0 / 82.0 &50.3 / 50.3 &45.7 / 45.5 & 24 \\
    NFGSM     &82.9 / 83.1 &47.3 / 47.2 &43.9 / 43.6 & 15 \\
    GAT       &81.9 / 81.9 &49.6 / 49.6 &45.2 / 45.1 & 33 \\
    GradAlign &80.8 / 80.8 &47.5 / 47.5& 43.0 / 43.0 & 53 \\
    PGI-MEP   &80.6 / 80.6 &51.0 / 51.0 &45.3 / 45.3 & 22 \\
    \rowcolor{black!10} Our CWR &81.1 / 81.1 &\textbf{\underline{52.4} / \underline{52.4}} &\textbf{\underline{46.1} / \underline{46.1}} & 24 \\
    \rowcolor{black!10} Our AGR &81.4 / 81.4 &\textbf{52.2 / 52.2} &\textbf{45.9 / 45.9} & 24 \\
    \bottomrule[1.5pt]
    \end{tabular}
    \label{CLRResults}
\end{table}

\begin{table}[t]
    \centering
    \caption{Performance Comparisons and Training Time Using WideResNet34-10 on the CIFAR-10. Bold Numbers Denote Better Than Previous Methods. Underline Represents the Best Robust Accuracy.}
    \setlength{\tabcolsep}{4mm}{
    \begin{tabular}{l | c|cccr}
    \toprule[1.5pt]
    Method &Clean &PGD-50 &AA &Time (h) \\
    \midrule
    FGSM-RS & 74.3 & 40.9 & 38.4 & 5.8 \\
    FGSM-SDI & 86.4 & 54.6 & 51.1 & 9.4 \\
    NFGSM & 80.6 & 47.9 & 44.6 & 5.8\\
    GAT & 85.2 & 54.9 & 50.0 & 12.9 \\
    GradAlign & 82.1 & 46.9 & 45.7 & 20.3 \\
    PGI-MEP & 85.1 & 56.4 & 51.4 & 8.3 \\
    \rowcolor{black!10} Our CWR &84.8 &\textbf{\underline{57.9}} &\textbf{\underline{52.7}} &8.4\\
    \rowcolor{black!10} Our AGR &85.2 &\textbf{57.1} &\textbf{52.1} &8.5\\
    \toprule[1.5pt]
    \end{tabular}
    }
    \label{WRNResults}
\end{table}

\subsection{Performance Comparison}
\subsubsection{Comparison Results on CIFAR-10} 
\begin{figure}[t]\centering
    \subfigure[FGSM-RS]{\includegraphics[scale=0.2]{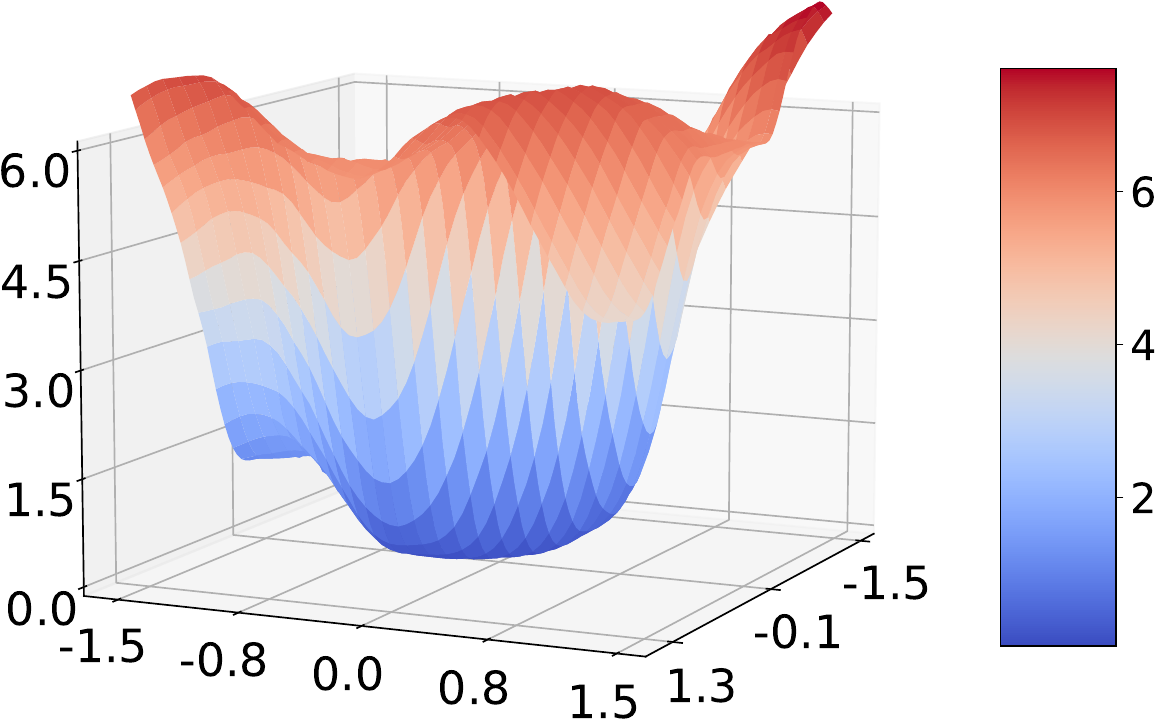}}
    \subfigure[GradAlign]{\includegraphics[scale=0.2]{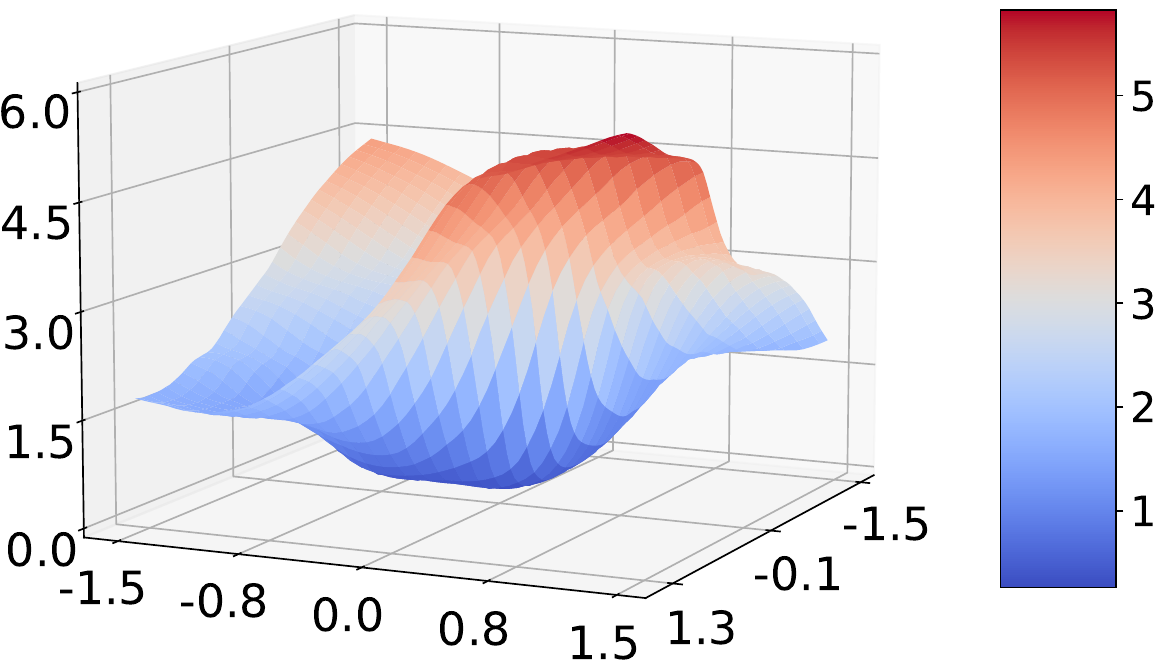}}
    \subfigure[PGI-MEP]{\includegraphics[scale=0.2]{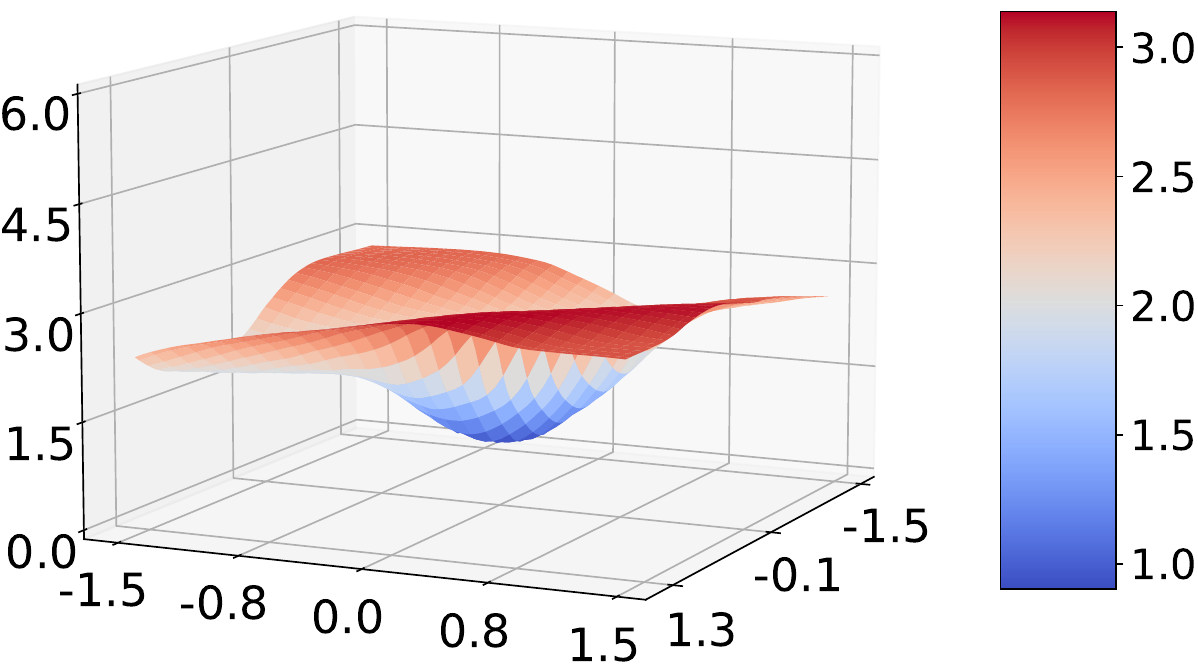}}
    \subfigure[FGSMSC]{\includegraphics[scale=0.2]{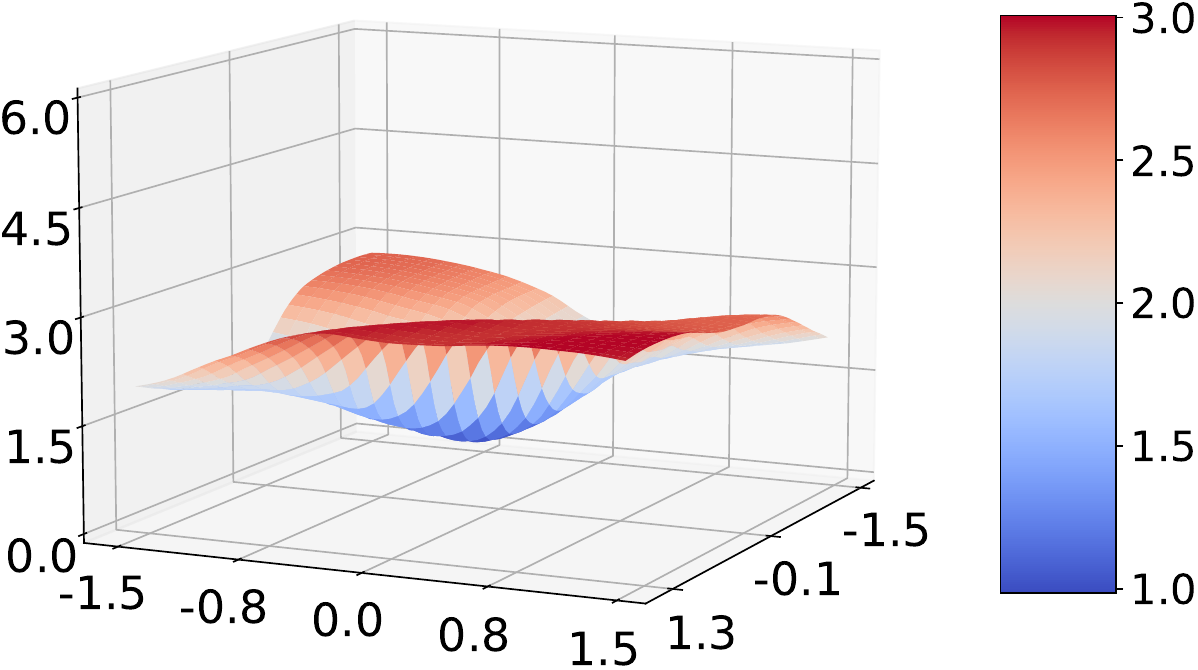}}
    \subfigure[CWR]{\includegraphics[scale=0.2]{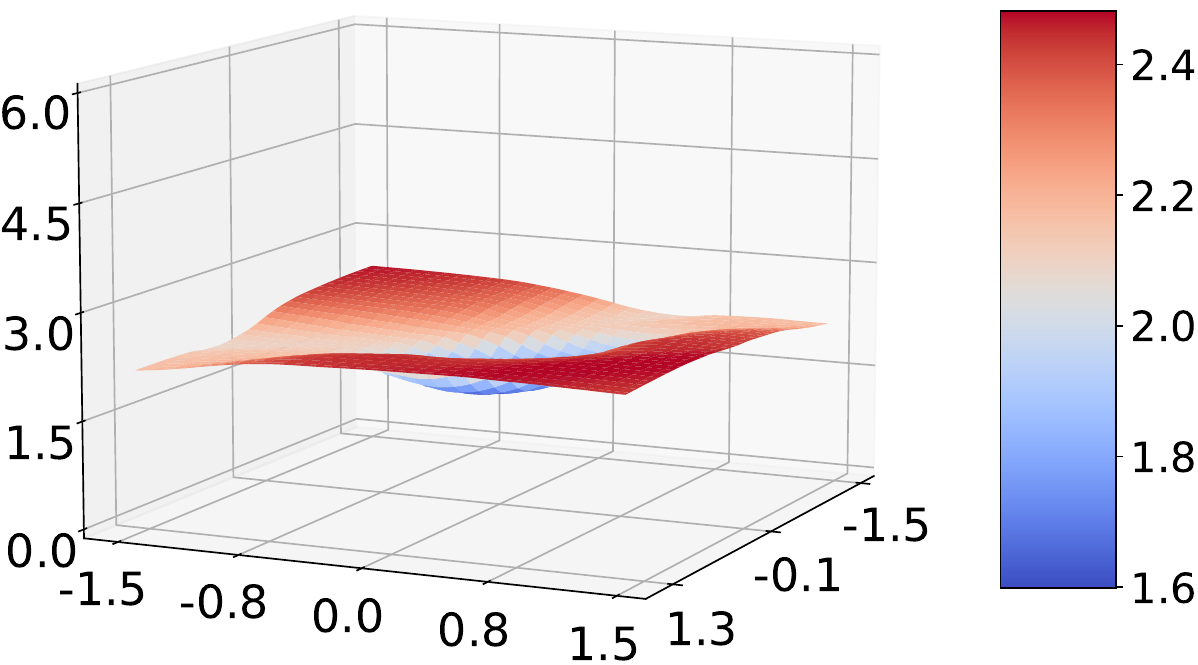}}
    \subfigure[AGR]{\includegraphics[scale=0.2]{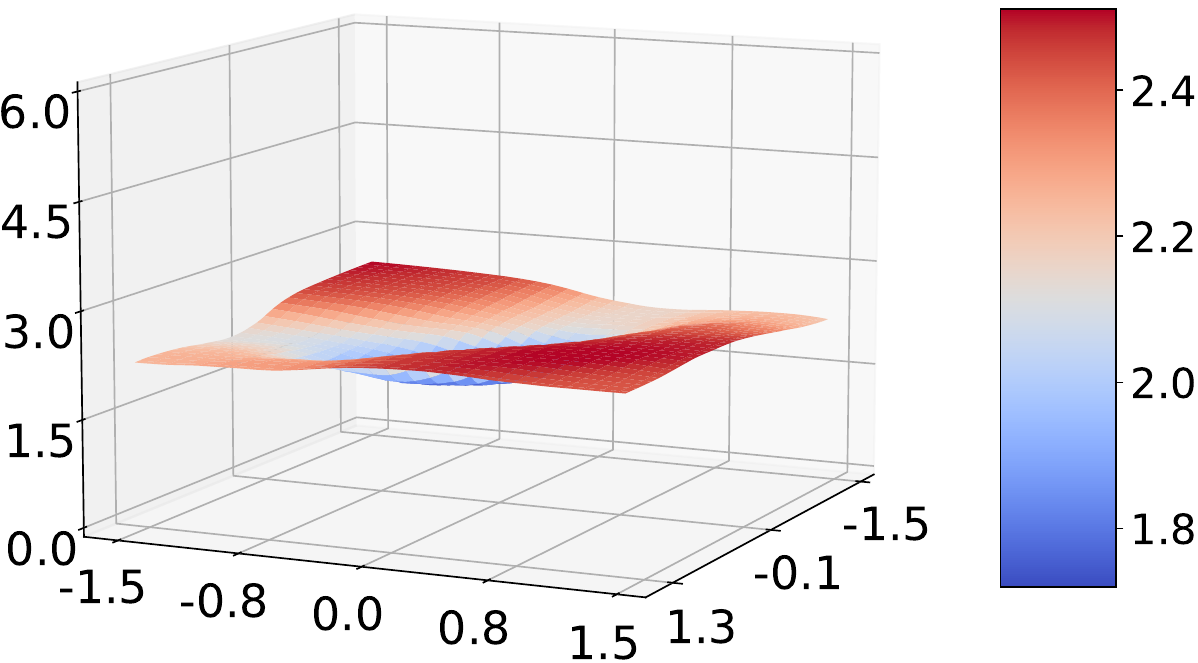}}
    \caption{Comparison of loss landscape on CIFAR100 dataset.}
    \label{LossSurCIFAR10}
\end{figure}

\begin{figure}[t]\centering
    \includegraphics[scale=0.35]{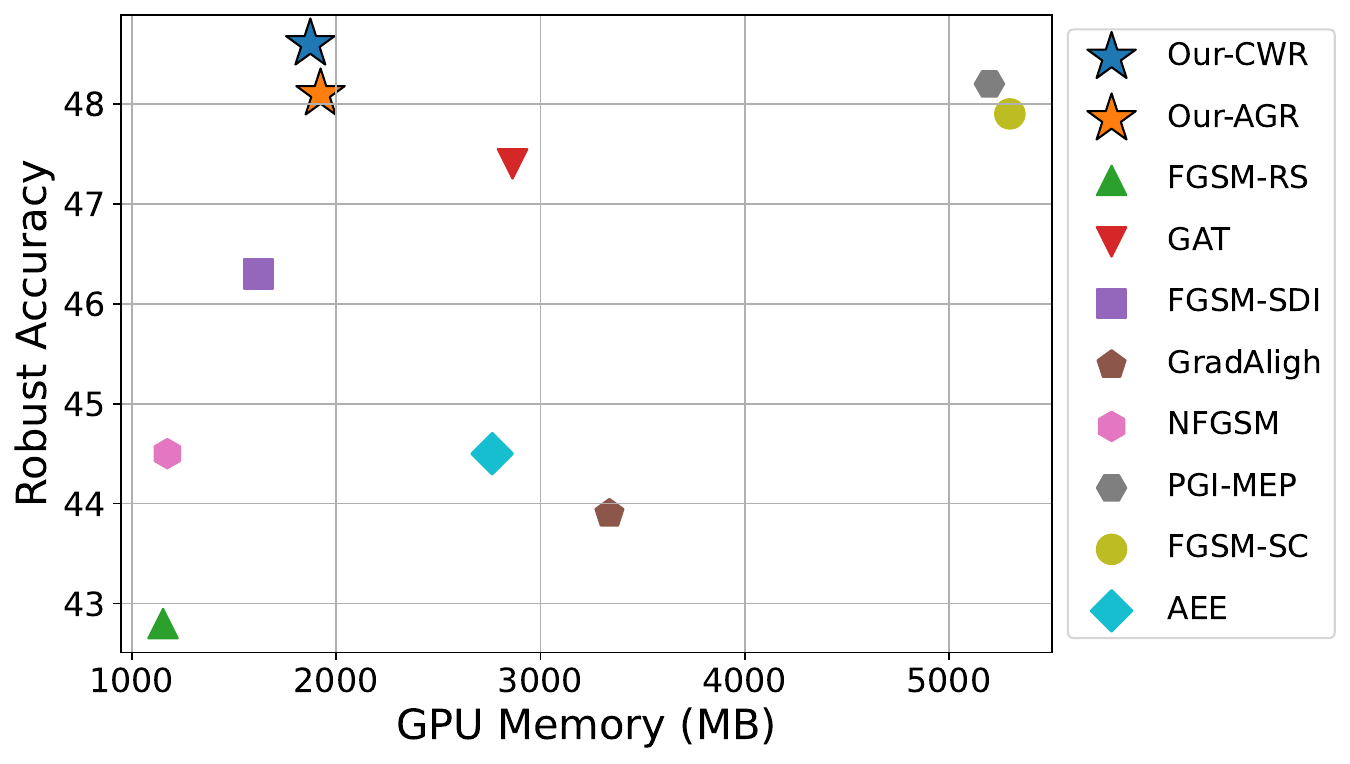}
    \caption{The GPU memory requirement and robustness under AutoAttack of various FAT methods using ResNet-18 on the CIFAR-10 dataset.}
    \label{GPUResults}
\end{figure}

The comparison of the best and final checkpoints on CIFAR-10 across different methods is presented in Table \ref{CIFARResults} (Left). The results demonstrate that our SKG-FAT with either CWR or AGR maintains competitive clean accuracy and achieves better robust accuracy compared to state-of-the-art methods, without suffering catastrophic overfitting. Specifically, SKG with CWR surpasses the state-of-the-art methods in defending against PGD-50 (+1.6\%), CW (+0.6\%), and AA (+0.4\%) at the best checkpoint, while achieving better clean accuracy. Moreover, our method significantly reduces computational time, achieving performance comparable to multi-step adversarial training with only one-fifth of the time cost. In terms of computational efficiency when compared with FAT methods, the SKG adds only slight overhead compared to FGSM-RS or NFGSM, as it only involves the inference of the model on clean examples and AEs. Nonetheless, our method significantly improves the robustness of the model, particularly against PGD-50, CW, and AA, with improvements of 6.5\%, 3.3\%, and 4.0\% at least, respectively. Furthermore, the comparison of GPU memory usage and robust accuracy against AA on the CIFAR-10 is presented in Fig. \ref{GPUResults}. The results demonstrate that our method achieves the best robust accuracy using less GPU memory. These findings underscore the superior efficiency of the proposed SKG-FAT compared to other methods, while also ensuring competitive computational efficiency. To ensure the reproducibility of the experimental results, we utilize three random seeds and report the average. The standard deviation of our CWR under AA is $\pm$0.28\%, while that of our AGR under AA is $\pm$0.23\%. We also visualize the loss landscape Fig. \ref{LossSurCIFAR10} for randomly sampled examples on the ResNet18, following the method in \cite{lossSurface}. Compared to other methods, the loss of our SKG-FAT with CWR or AGR demonstrates greater linearity in the adversarial direction. This verifies that incorporating our SKG-FAT can better preserve the local linearity of the model, which is the reason the SKG-FAT can achieve better adversarial robustness.

\begin{table*}[t]
	\centering
	\caption{Accuracy Comparisons on Tiny Imagenet and Imagenet-100. Bold Numbers Denote Better Than Previous Methods. Underline Represents the Best Robust Accuracy.}
	\setlength{\tabcolsep}{1.7mm}{
	\begin{tabular}[l]{@{}l| c c c c c| c c c c c}
	\toprule[2pt]
	\multirow{3}*{Method} &\multicolumn{5}{c|}{Tiny ImageNet} &\multicolumn{5}{c}{ImageNet-100}\\
    \cmidrule(r){2-6} \cmidrule(r){7-11}
	&Clean &MI &PGD50 &CW &AA &Clean &MI &PGD50 &CW &AA\\
    &Best / Last &Best / Last &Best / Last &Best / Last &Best / Last &Best / Last &Best / Last &Best / Last &Best / Last  &Best / Last\\
	\toprule[1pt]
    \multirow{1}*{FGSM-RS}   &43.5 / 43.5 &17.8 / 17.8 &16.8 / 16.8 &14.8 /14.8 &13.3 / 13.3 &54.6 / 54.6 &22.6 / 22.6 &25.3 / 25.3 &23.8 / 23.8 &19.2 / 19.2\\
	\multirow{1}*{GAT}       &46.0 / 46.3 &15.6 / 14.8 &14.3 / 13.6 &12.7 / 12.2 &10.2 / 9.7 &66.4 / 67.0 &36.6 / 36.4 &36.4 / 36.2 &32.4 / 32.3 &28.4 / 28.1\\
	\multirow{1}*{FGSM-SDI}  &44.1 / 45.4 &20.9 / 17.6 &20.2 / 16.7 &17.3 / 14.8 &15.5 / 12.4 &62.4 / 62.4 &37.9 / 37.9 &37.2 / 37.1 &32.4 / 32.4 &28.8 / 28.7\\
	\multirow{1}*{GradAlign} &38.2 / 38.3 &17.4 / 17.4 &16.8 / 16.7 &14.1 / 13.9 &12.6 / 12.4 &-    &-    &-    &-    &-   \\
	\multirow{1}*{N-FGSM}    &46.1 / 46.1 &17.2 / 17.2 &16.0 / 16.0 &14.9 / 14.9 &12.7 / 12.7 &61.5 / 61.5 &35.3 / 35.3 &33.5 / 33.5 &31.0 / 31.0 &27.7 / 27.7\\
    \multirow{1}*{PGI-BP}    &45.1 / 45.3 &22.8 / 21.6 &21.4 / 20.2 &17.4 / 16.2 &15.3 / 14.4 &63.5 / 63.8 &36.5 / 36.1 &38.0 / 38.0 &33.1 / 32.9 &28.8 / 28.6\\
    \multirow{1}*{PGI-MEP}   &44.6 / 45.1 &23.9 / 22.3 &23.0 / 21.9 &17.8 / 16.8 &16.6 / 15.9 &63.2 / 63.5 &36.8 / 36.4 &38.3 / 38.0 &33.3 / 33.0 &29.2 / 29.0\\
	\multirow{1}*{AEE}       &48.5 / 49.0 &21.8 / 21.0 &18.1 / 17.3 &17.4 / 17.0 &15.2 / 14.4 &-    &-    &-    &-    &- \\
 
	\rowcolor{black!10}\multirow{1}*{\textbf{Our CWR}} &45.9 / 47.5 &\textbf{\underline{25.1} / \underline{24.2}} &\textbf{\underline{24.4} / \underline{23.2}} &\textbf{\underline{19.4} / \underline{18.1}} &\textbf{\underline{17.5} / \underline{16.3}} &65.8 / 66.0 &\textbf{\underline{39.0} / \underline{39.0}} &\textbf{\underline{39.9} / \underline{39.8}} &\textbf{\underline{34.2} / \underline{34.0}} &\textbf{\underline{30.1} / \underline{29.8}}\\
	
    \rowcolor{black!10}\multirow{1}*{\textbf{Our AGR}} &45.5 / 47.2 &\textbf{24.8 / 24.0} &\textbf{23.9 / 22.8} &\textbf{18.8 / 17.5} &\textbf{16.9 / 16.1} &64.7 / 65.4 &\textbf{38.2 / 38.2} &\textbf{38.8 / 38.7} &\textbf{33.7 / 33.5} &29.1 / 28.8 \\
	\toprule[1pt]
	\multicolumn{11}{c}{\qquad \qquad \qquad Multi-step Adversarial Training}\\
	\toprule[1pt]
	\multirow{1}*{LAS-AWP} &47.1 / 47.8 &24.9 / 24.5 &23.6 / 23.6 &20.4 / 20.4 &18.5 / 18.5 &64.5 / 64.5 &38.8 / 38.8 &32.3 / 40.1 &35.4/ 35.4 &32.4 / 32.4\\
	\multirow{1}*{MART}    &38.7 / 36.8 &21.1 / 12.7 &20.6 / 11.9 &10.3 / 16.8 &15.6 / 9.6  &64.8 / 64.8 &39.9 / 38.7 &37.9 / 37.9 &35.1 / 35.1 &31.8 / 31.8\\
	\toprule[2pt]
\end{tabular}
}
\label{INResults}
\end{table*}

To evaluate the performance of our SKG-FAT under different learning rate strategy, we adopt a cyclic learning rate strategy with 40 training epochs for the experiments. The corresponding results are presented in Table \ref{CLRResults}. Our SKG-FAT also performs well in this setting. The SKG-FAT with CWR achieved improvements of 1.4\% and 0.7\% against CW and AA at the best checkpoint, respectively. Meanwhile, the SKG-FAT with AGR achieved improvements of 1.1\% and 0.5\% against CW and AA at the best checkpoint, respectively. These results verify that the robustness enhancement provided by our method is a substantial improvement in training effectiveness, rather than merely a result of the training setting.
\par
The results using WideResNet34-10 \cite{WRN} to evaluate different methods on the CIFAR-10 are provided in Table \ref{WRNResults}. Due to the significantly larger scale of Wide ResNet-34 compared to ResNet-18, all methods exhibit improved accuracy over the results obtained with ResNet-18. However, our method still achieves better robust accuracy compared to other methods, with improvements of 1.5\% and 1.3\% against PGD-50 and AA, respectively. This confirms the scalability of our approach, demonstrating its ability to enhance robust accuracy across different models, rather than being limited to improving the robustness of the single model.
\par

\subsubsection{Comparison Results on CIFAR-100} The comparison results on CIFAR-100 are presented in Table \ref{CIFARResults} (Right). The corresponding results are consistent with those observed on CIFAR-10, where our method achieves the best adversarial robustness compared to existing FAT methods, while also preventing catastrophic overfitting. Specifically, our approach with CWR enhances the robustness when defending PGD-50 for +1.0\%, CW for +0.6\%, and AA for +0.4\%. These results suggest that our FAT with CWR or AGR can effectively improve the adversarial robustness of the model. In terms of training efficiency, our method continues to hold an advantage, achieving the best robust accuracy with reduced computational time. Notably, while the time cost of SKG-FAT is slightly higher than that of FGSM-RS and NFGSM, it results in significant improvements, enhancing defense against PGD-50, CW, and AA attacks by 5.5\%, 2.5\% and 3.1\% at least, respectively. Moreover, to ensure the reproducibility of the experimental results, we utilize three random seeds and report the average. The standard deviation of our CWR under AA is $\pm$0.28\%, while that of our AGR under AA is $\pm$0.23\%. These findings demonstrate that the proposed SKG-FAT achieves superior robustness on datasets with a larger number of classes.

\begin{figure}[t]\centering
	\includegraphics[scale=0.43]{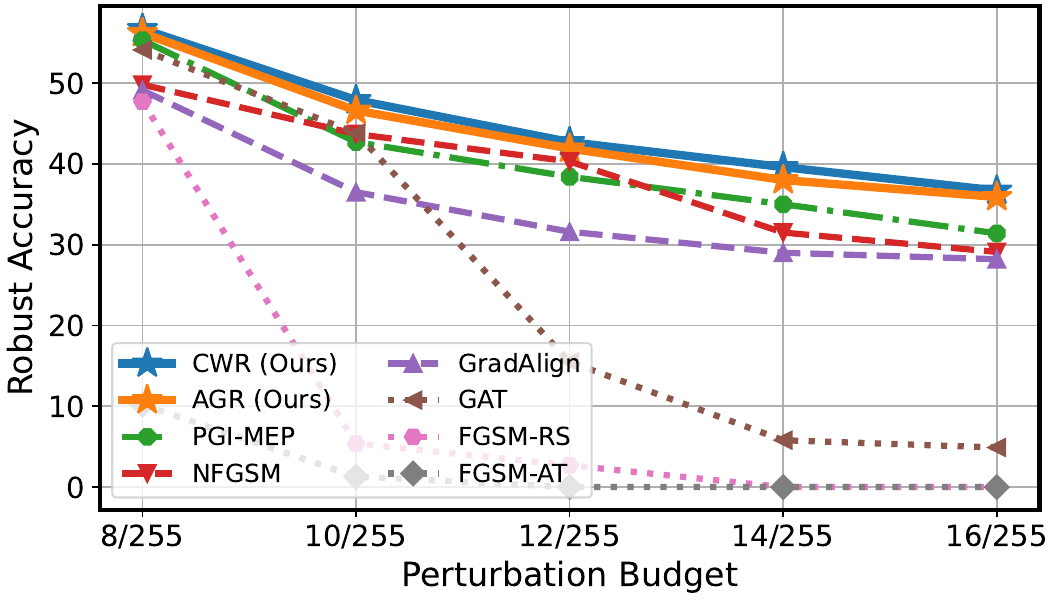}
	\caption{Comparison of different methods trained on ResNet-18 with varying perturbation budgets on the CIFAR-10 when defending against PGD-10.}
	\label{DiffEps}
\end{figure}

\begin{figure*}[t]\centering
    \includegraphics[scale=0.29]{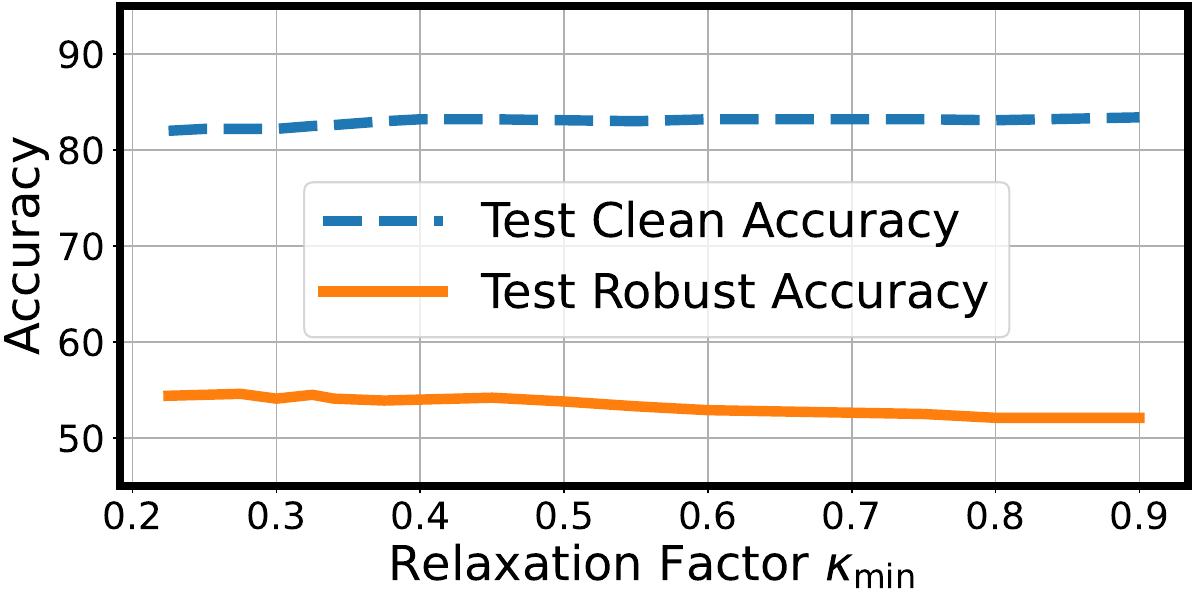}
    \includegraphics[scale=0.29]{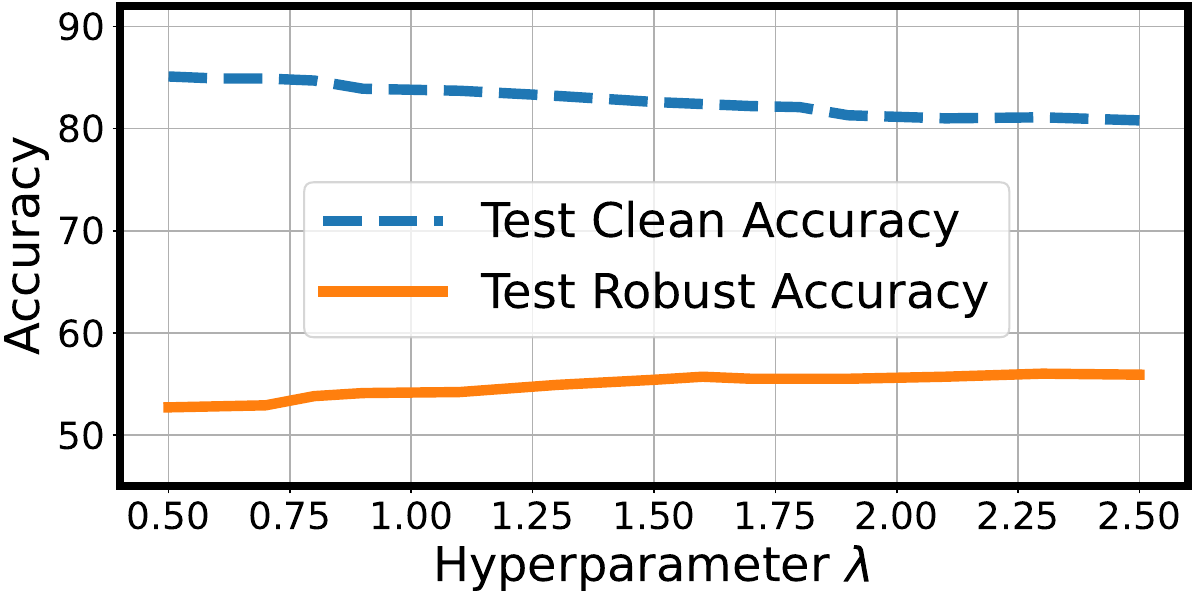}
    \includegraphics[scale=0.29]{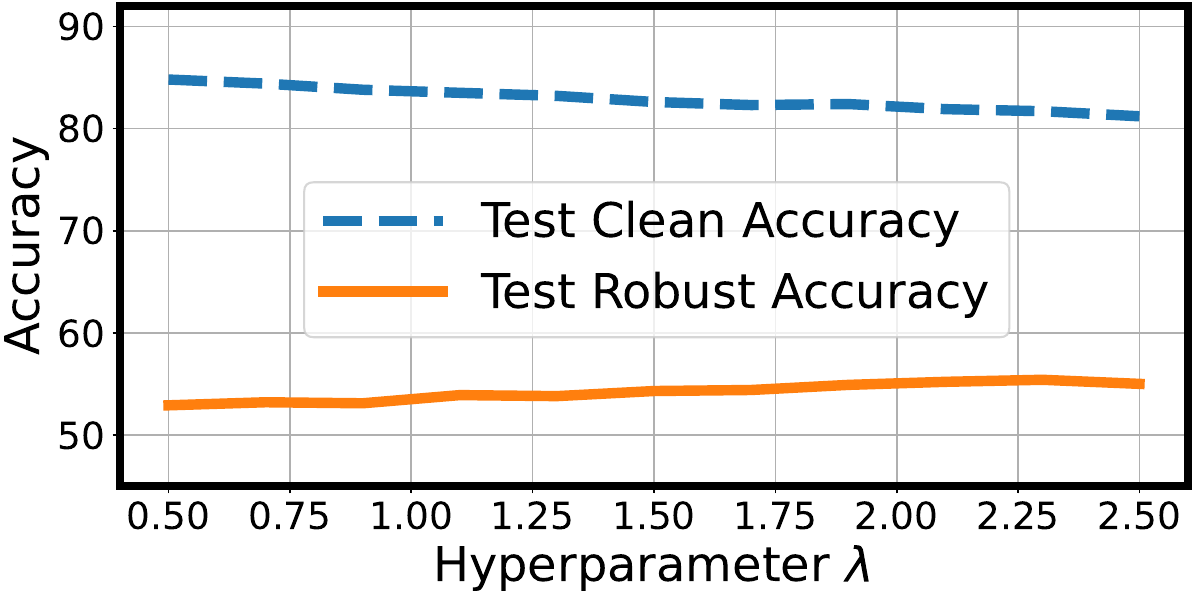}
    \caption{Effect of hyperparameters variation on SKG-FAT. The results are evaluated by PGD-10 attack. Left Minimum relaxation factor $\kappa_{\text{min}}$ in equation \eqref{LER}. Middle: Hyperparameter $\lambda$ in CWR \eqref{CWRFormula}. Right: Hyperparameter $\lambda$ in AGR \eqref{AGRFormula}.}
    \label{HyperAblation}
\end{figure*}

\subsubsection{Comparison Results on Tiny ImageNet} The comparison results on the Tiny ImageNet dataset are provided in Table \ref{INResults} (Left). This dataset contains more classes and higher resolution than the CIFAR-10/100 datasets. For Tiny ImageNet, our method demonstrates stronger competitiveness. Specifically, while maintaining similar clean accuracy, our SKG-FAT with CWR or AGR achieves significant improvements in adversarial robustness. For SKG-FAT with AGR, significant improvements are observed against PGD-50, CW, and AA attacks, while SKG-FAT with CWR further surpasses the state-of-the-art methods, achieving a 1.4\%, 1.6\%, and 0.9\% improvement in robustness against CW and AA, respectively. Overall, the proposed SKG-FAT with CWR or AGR demonstrates more pronounced advantages on datasets like Tiny ImageNet with a larger number of classes. This is because our method adjusts the training configuration for each class based on class-wise performance, leading to more significant improvements in adversarial robustness as the number of classes increases. 

\subsubsection{Comparison Results on ImageNet-100} The comparison results on ImageNet-100 are presented in Table \ref{INResults} (Right). The results indicate that our SKG-FAT remains competitive even on this higher-resolution dataset. Specifically, our SKG-FAT with CWR achieves the best robust accuracy, surpassing the state-of-the-art methods by 0.8\% and 0.9\% in defense against CW and AA at the best checkpoint, respectively. Notably, the GPU memory requirement of our SKG-FAT with CWR or AGR is only dependent on the batch size, allowing for single-GPU training even on large datasets like ImageNet-100. In contrast, the previous best method, PGI-MEP, requires loading the entire dataset simultaneously \cite{PGIFGSM}, leading to a significant increase in GPU memory usage as the dataset size grows. These findings confirm that the SKG-FAT can achieve state-of-the-art adversarial robustness on large-scale datasets.

\begin{figure}[t]\centering
	\includegraphics[scale=0.21]{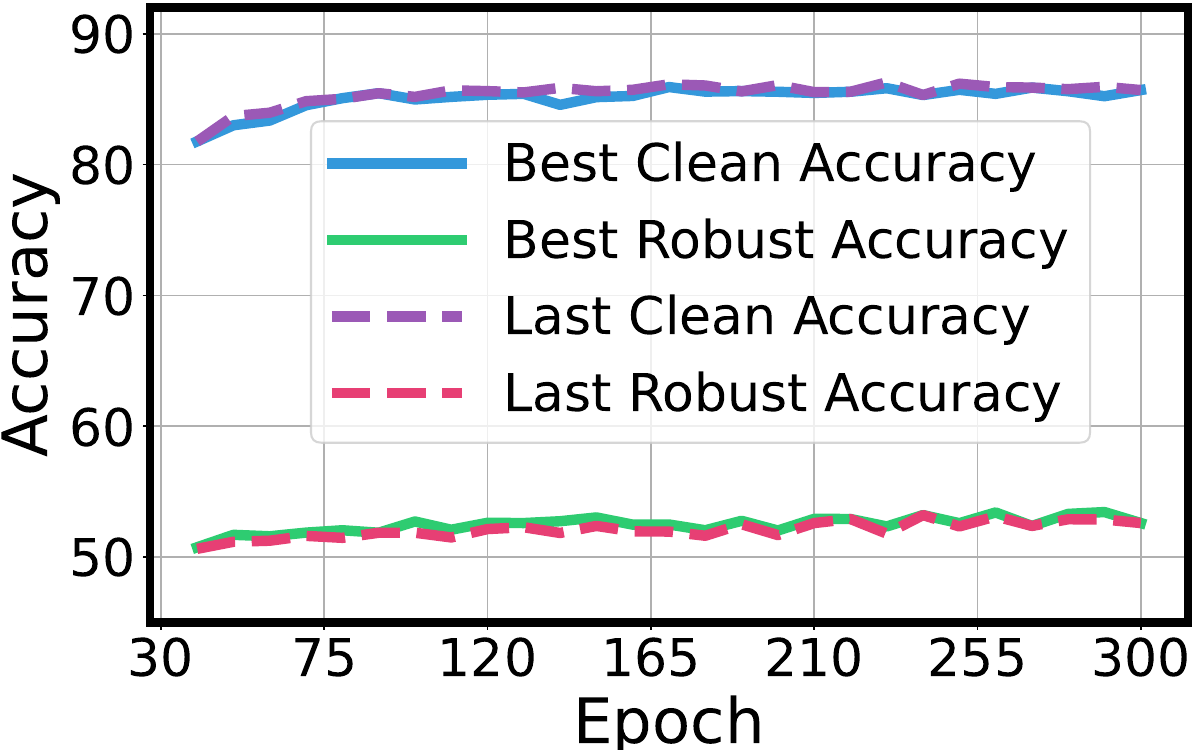}
	\includegraphics[scale=0.21]{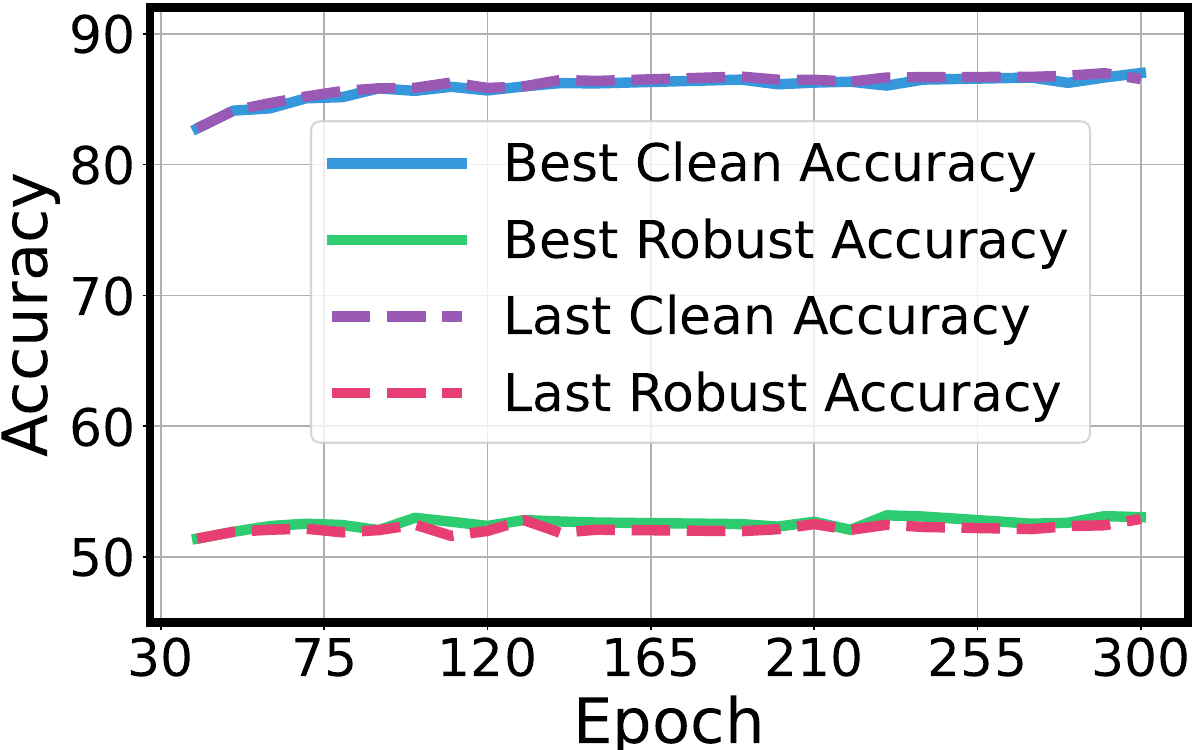}
	\caption{Best and last accuracy of SKG-FAT, which adopts different total training epochs across 40 to 300 with the interval of 10 epochs. The experiments are performed on ResNet18 with the CIFAR-10. Left: SKG-FAT with AGR. Right: SKG-FAT with CWR.}
	\label{DiffEpoch}
\end{figure}

\subsubsection{Different Perturbation Budgets}
Catastrophic overfitting is a critical issue in the FAT field, as it leads to the collapse of model robustness and undermines the effectiveness of adversarial training \cite{UCOSS}. To validate that our method can overcome catastrophic overfitting and maintain competitiveness under various perturbation budgets, we employ the settings from \cite{PGIFGSM} and compare our approach with other FAT methods on CIFAR-10 using ResNet18 as the backbone. This comparison investigates performance across different perturbation budgets. The robust accuracy is evaluated by PGD-10, and the results are presented in Fig. \ref{DiffEps}. Previous FAT methods suffer when training with larger perturbation budgets (as in FGSM-AT, FGSM-RS, and GAT). Conversely, our proposed SKG-FAT can eliminate catastrophic overfitting and improve robust accuracy across different perturbation budgets. This is because our SKLR can alleviate the imbalance between inner and outer optimization in the minimax problem of FAT, thereby stabilizing training and overcoming catastrophic overfitting. Additionally, this method can adjust class-wise based on the training status to further enhance robustness.

\subsection{Effect of Hyperparameters.} 
The effect of hyperparameters on the CIFAR-100 is presented in Fig. \ref{HyperAblation}. As shown in Fig. \ref{HyperAblation} (Left), the factor $\kappa_{\min}$ in SKLR \eqref{LER} varies from $0.225$ to $0.9$ for the evaluation, sampled $15$ times within this interval. First, the factor $\kappa_{\min}$ shows minimal impact on clean accuracy, with only a $1.3\%$ difference between the highest and lowest values. Conversely, the factor $\alpha_{\min}$ is inversely proportional to robust accuracy, with the highest and lowest gap reaching $1.4\%$ within the interval. The ablation study on the hyperparameter $\lambda$ for CWR and AGR is presented in Figs. \ref{HyperAblation} (Middle) and (Right). Specifically, the hyperparameter $\lambda$ is directly proportional to robust accuracy and inversely proportional to clean accuracy. These results highlight the stability of the regularization in the SKG-FAT. As parameter lambda varies, the robust accuracy difference for CWR is $2.2\%$, while AGR achieves $2.1\%$.

\begin{table}[t]
    \centering
    \caption{Ablation study of SKG-FAT on CIFAR-100 dataset. Reg$^\dagger$ represents self-knowledge guided regularization. Bold numbers denote the best robust accuracy.}
    \setlength{\tabcolsep}{0.9mm}{
    \begin{tabular}[l]{@{}c| c c| c c c c| c c c c}
    \toprule[2pt]
    &\multirow{2}*{Reg$^\dagger$} &\multirow{2}*{SKLR} &\multicolumn{4}{c|}{Best Epoch} &\multicolumn{4}{c}{Last Epoch}\\
    && &Clean &PGD10 &CW &AA &Clean &PGD10 &CW &AA\\
    \toprule[1pt]
    \multirow{4}*{CWR}&\xmark &\xmark &54.6 &24.5 &22.9 &20.7 &54.9 &24.5 &22.9 &20.7\\
    &\xmark &\cmark &58.0 &27.7 &24.3 &22.2 &59.2 &28.3 &24.1 &22.0\\
    &\cmark &\xmark &56.5 &30.9 &25.5 &23.0 &58.9 &30.3 &25.3 &22.6\\
    &\cmark &\cmark &57.8 &\textbf{32.0} &\textbf{27.8} &\textbf{25.4} &57.9 &\textbf{31.9} &\textbf{27.5} &\textbf{25.3}\\
    \toprule[1pt]
    \multirow{2}*{AGR}&\cmark &\xmark &57.5 &30.9 &25.5 &23.0 &58.1 &30.3 &25.3 &22.6 \\
    &\cmark &\cmark &57.5 &\textbf{31.9} &\textbf{27.6} &\textbf{25.5} &57.9 &\textbf{31.6} &\textbf{27.4} &\textbf{25.2}\\
    \toprule[2pt]
\end{tabular}
}
\label{Ablation}
\end{table}

\subsection{Ablation Study.}
Ablation results assessing the impact of each component of SKG-FAT on CIFAR-100 are presented in Table \ref{Ablation}. The reported metrics include clean and robust accuracy under different attacks for both the best and last epochs. FGSM-RS is employed as the benchmark for experiments. Specifically, both CWR and AGR enhance the clean and robust accuracy. Meanwhile, SKLR further improves the clean accuracy, albeit with a slight degradation in robust accuracy. Note that the clean and robust accuracy achieved in the last training epoch of the SKG-FAT is equal to the accuracy obtained in the best epoch (or shows only slight differences). This indicates that our method exhibits competitive training stability, eliminating additional early stop operations to trade off optimal results.

\subsection{Analyses on Training Epochs}
Catastrophic overfitting poses a challenge for FAT, significantly reducing the robustness of the model against multi-step adversarial attacks \cite{FATSC}. To further verify the effectiveness of our SKG-FAT in addressing the catastrophic overfitting, we perform SKG-FAT with CWR and AGR on the CIFAR-10 with epochs taken every 10 in the range of 40 to 300 as shown in Fig. \ref{DiffEpoch}. The results demonstrate that our method achieves steady clean and robust accuracy under different total training epochs. This indicates that our method consistently maintains training stability even with additional training epochs, while also further improving the adversarial robustness of the model as training progresses. Moreover, the minimal difference between the last and best checkpoints, even after training with more epochs, demonstrates that our method effectively relieves the requirement to balance or trade-off between checkpoints.

\begin{figure}[t]\centering
    \includegraphics[scale=0.26]{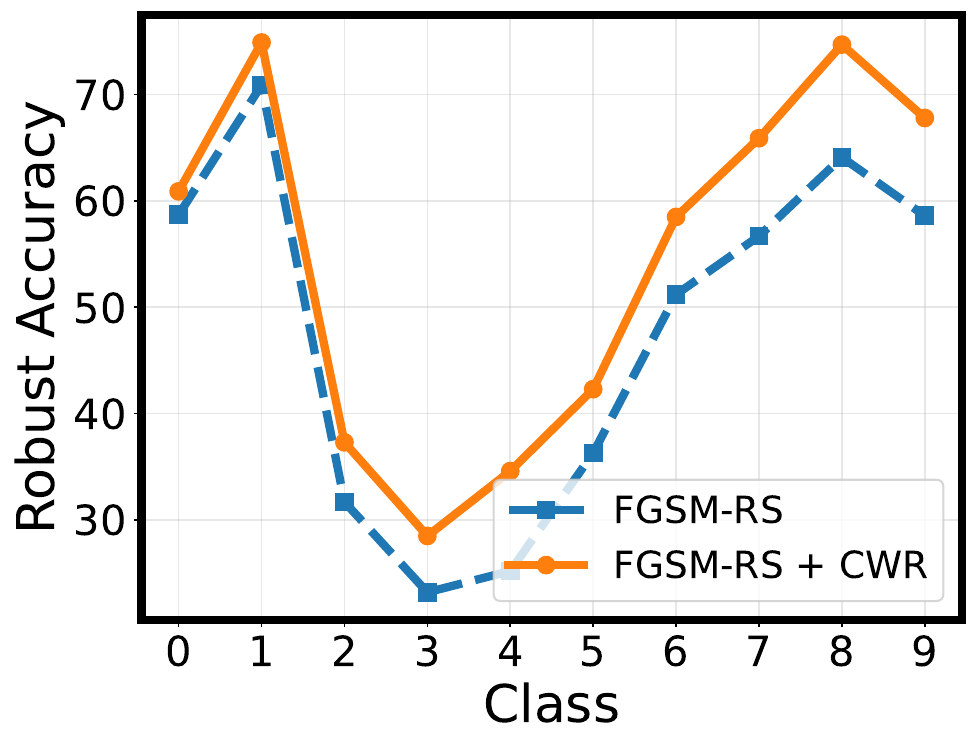}
    \includegraphics[scale=0.26]{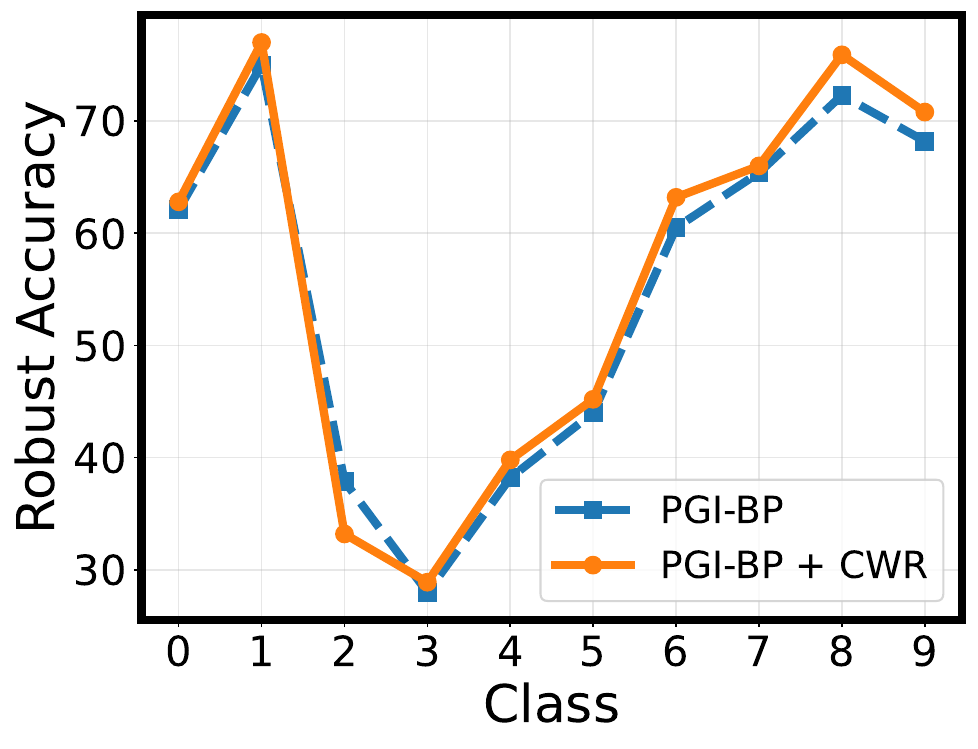}
    \caption{Comparison of class-wise accuracy on CIFAR-10 in case of with or w/o CWR. Left: Results with FGSM-RS. Right: Results with PGI-BP.}
    \label{PlugCWR}
\end{figure}
\subsection{Discussion of Reasons for SKG-FAT Improve Robustness}
To understand the underlying mechanism of how our method alleviates accuracy misalignment and disparity to improve model robustness, we integrate self-knowledge guided regularization and label relaxation as plug-in components into the baseline methods PGI-BP and FGSM-RS (consistent with our analysis in Section \ref{PAs}). We then compare the performance differences between these enhanced methods and the baselines and the results are presented in Fig \ref{PlugCWR}. As shown in Fig. \ref{PlugCWR}, our CWR assigns differentiated regularization weights to each class based on its accuracy disparity. This approach enhances the robust accuracy of high-accuracy classes while preserving the robust accuracy of lower-accuracy classes. The resulting improvement in class-wise accuracy contributes to an overall enhancement of the model’s adversarial robustness.

\begin{figure}[t]\centering
    \includegraphics[scale=0.265]{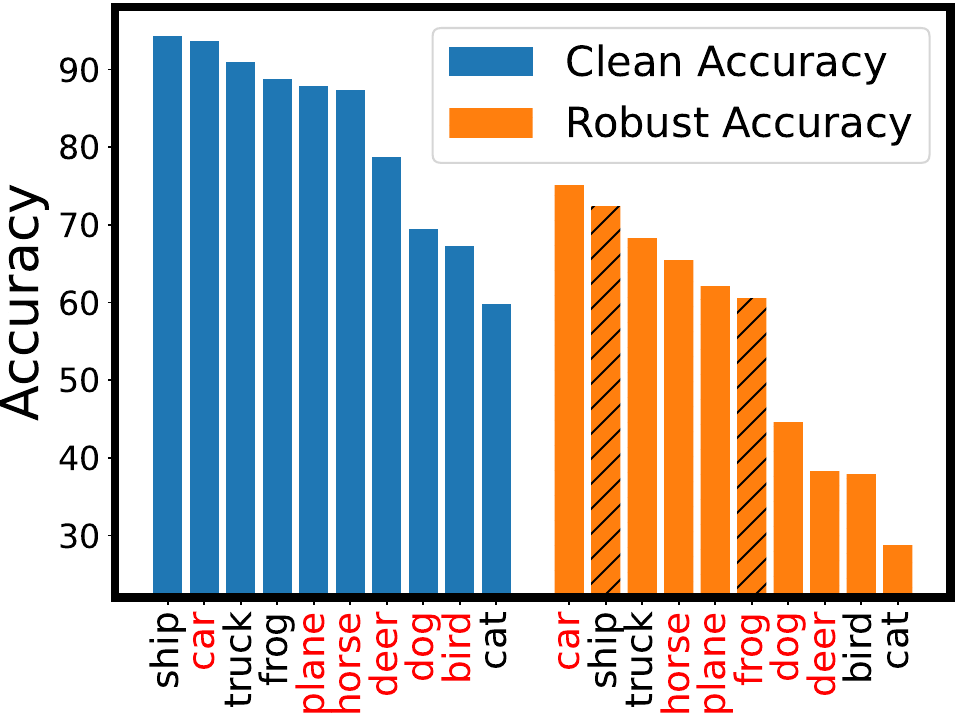}
    \includegraphics[scale=0.265]{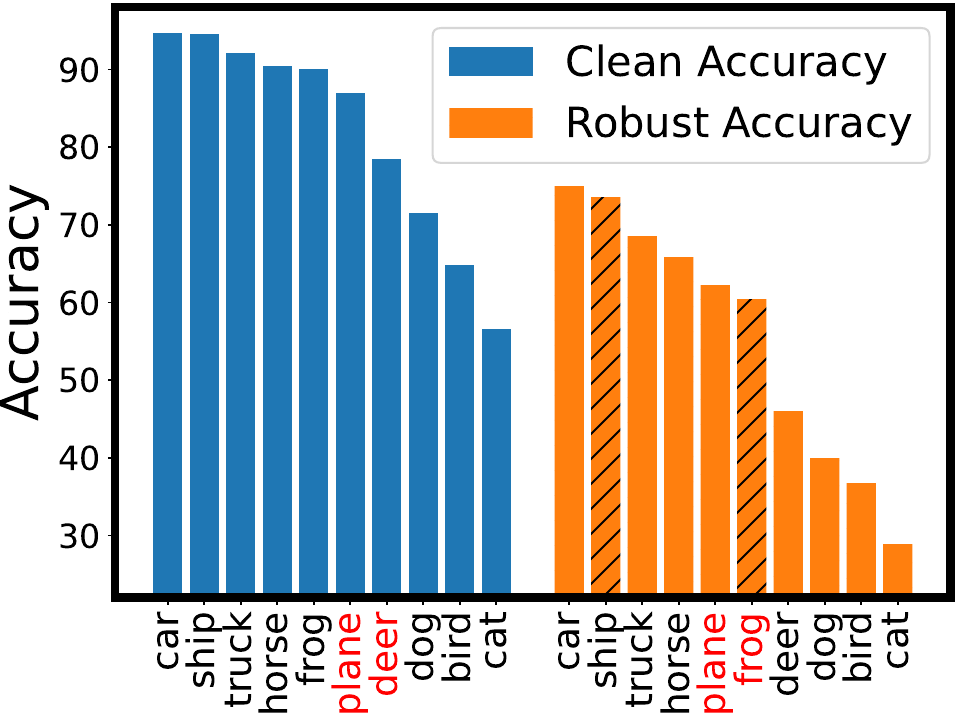}
    \caption{Comparison of class-wise accuracy and accuracy misalignment on CIFAR-10 in case of with or w/o AGR. Red highlights mean that the class accuracy is misalignment. Left: Results with vanilla PGI-BP. Right: Results with AGR enhanced PGI-BP.}
    \label{PlugAGR}
\end{figure}
In subsection \ref{FFWP}, our analysis identified classes with misalignment between clean and robust accuracy (GCBR and BCGR). Based on this observation, we designed AGR to address this misalignment by emphasizing regularization strength for these misaligned class examples, thereby improving model robustness. To validate the mechanism behind AGR, we integrate it as a plug-in to enhance PGI-BP and compare its class-wise accuracy with the baseline method. Experiments were conducted using the CIFAR-10 dataset with PGD-10 for testing. As the results shown in Fig. \ref{PlugAGR}, reveal that vanilla PGI-BP exhibits misalignment in six classes. However, with AGR integrated as a plug-in, this misalignment is reduced to just two classes with only a 1.8\% robust accuracy gap. This reduction indicates that AGR effectively alleviates clean and robust accuracy misalignment, which contributes to the improvement in model robustness.

\section{Conclusion and Limitation}\label{CFSec}
We analyzed the accuracy of the model on different examples during training to investigate how differences in example clean and robust accuracy affect fast adversarial training (FAT). This investigation is performed from the class-wise and our proposed accuracy alignment perspective. Corresponding results indicate disparities in clean and robust accuracy across examples with distinct classes. Some examples showed no positive correlation between clean and robust accuracy. Based on these observations, we have proposed the self-knowledge guided (SKG) FAT, which requires only one additional hyperparameter with a defined range for regularization-involved methods. Comprehensive experiments demonstrate that the SKG-FAT can improve the robustness of models without compromising training efficiency. Despite conducting extensive evaluations of FAT using the proposed perspective of accuracy alignment, our testing was primarily conducted on two representative methods due to computational constraints. Thus, further investigations into different methods may uncover patterns that extend beyond the scope of our discussion. We hope that future research will address these limitations.

\bibliographystyle{IEEEtran}
\bibliography{Manuscript}{}

\begin{thebibliography}{10}
\providecommand{\url}[1]{#1}
\csname url@samestyle\endcsname
\providecommand{\newblock}{\relax}
\providecommand{\bibinfo}[2]{#2}
\providecommand{\BIBentrySTDinterwordspacing}{\spaceskip=0pt\relax}
\providecommand{\BIBentryALTinterwordstretchfactor}{4}
\providecommand{\BIBentryALTinterwordspacing}{\spaceskip=\fontdimen2\font plus
\BIBentryALTinterwordstretchfactor\fontdimen3\font minus
  \fontdimen4\font\relax}
\providecommand{\BIBforeignlanguage}[2]{{%
\expandafter\ifx\csname l@#1\endcsname\relax
\typeout{** WARNING: IEEEtran.bst: No hyphenation pattern has been}%
\typeout{** loaded for the language `#1'. Using the pattern for}%
\typeout{** the default language instead.}%
\else
\language=\csname l@#1\endcsname
\fi
#2}}
\providecommand{\BIBdecl}{\relax}
\BIBdecl

\bibitem{han2022survey}
K.~Han, Y.~Wang, H.~Chen, X.~Chen, J.~Guo, Z.~Liu, Y.~Tang, A.~Xiao, C.~Xu,
  Y.~Xu \emph{et~al.}, ``A survey on vision transformer,'' \emph{IEEE Trans.
  Pattern Anal. Mach. Intell.}, vol.~45, no.~1, pp. 87--110, 2022.

\bibitem{mao2022towards}
X.~Mao, G.~Qi, Y.~Chen, X.~Li, R.~Duan, S.~Ye, Y.~He, and H.~Xue, ``Towards
  robust vision transformer,'' in \emph{CVPR}, 2022, pp. 12\,042--12\,051.

\bibitem{RLFA}
Q.~Xu, Z.~Yang, Y.~Zhao, X.~Cao, and Q.~Huang, ``Rethinking label flipping
  attack: From sample masking to sample thresholding,'' \emph{IEEE Trans.
  Pattern Anal. Mach. Intell.}, 2023.

\bibitem{FASTEN}
L.~Huang, Q.~Huang, P.~Qiu, S.~Wei, and C.~Gao, ``Fasten: Fast ensemble
  learning for improved adversarial robustness,'' \emph{IEEE Trans. Inf.
  Forensics Secur.}, vol.~19, pp. 2565--2580, 2024.

\bibitem{chen2023adaptive}
B.~Chen, J.~Yin, S.~Chen, B.~Chen, and X.~Liu, ``An adaptive model ensemble
  adversarial attack for boosting adversarial transferability,'' in
  \emph{ICCV}, 2023, pp. 4489--4498.

\bibitem{10504304}
R.~Liu, W.~Zhou, T.~Zhang, K.~Chen, J.~Zhao, and K.-Y. Lam, ``Boosting
  black-box attack to deep neural networks with conditional diffusion models,''
  \emph{IEEE Trans. Inf. Forensics Secur.}, vol.~19, pp. 5207--5219, 2024.

\bibitem{AADGD}
L.~Sun, Y.~Dou, C.~Yang, K.~Zhang, J.~Wang, S.~Y. Philip, L.~He, and B.~Li,
  ``Adversarial attack and defense on graph data: A survey,'' \emph{IEEE Trans.
  Knowledge Data Eng.}, 2022.

\bibitem{AEANRF}
A.~Li, Y.~Wang, Y.~Guo, and Y.~Wang, ``Adversarial examples are not real
  features,'' in \emph{NeurIPS}, 2023.

\bibitem{huang2024pointcat}
Q.~Huang, X.~Dong, D.~Chen, H.~Zhou, W.~Zhang, K.~Zhang, G.~Hua, Y.~Cheng, and
  N.~Yu, ``Pointcat: Contrastive adversarial training for robust point cloud
  recognition,'' \emph{IEEE Trans. Image Process.}, 2024.

\bibitem{FATASS}
Z.~Huang, Y.~Fan, C.~Liu, W.~Zhang, Y.~Zhang, M.~Salzmann, S.~S{\"u}sstrunk,
  and J.~Wang, ``Fast adversarial training with adaptive step size,''
  \emph{IEEE Trans. Image Process.}, 2023.

\bibitem{FSRAR}
W.~J. Kim, Y.~Cho, J.~Jung, and S.-E. Yoon, ``Feature separation and
  recalibration for adversarial robustness,'' in \emph{CVPR}, 2023, pp.
  8183--8192.

\bibitem{TopAlignAT}
H.~Kuang, H.~Liu, X.~Lin, and R.~Ji, ``Defense against adversarial attacks
  using topology aligning adversarial training,'' \emph{IEEE Trans. Inf.
  Forensics Secur.}, vol.~19, pp. 3659--3673, 2024.

\bibitem{MutAT}
J.~Liu, C.~P. Lau, H.~Souri, S.~Feizi, and R.~Chellappa, ``Mutual adversarial
  training: Learning together is better than going alone,'' \emph{IEEE Trans.
  Inf. Forensics Secur.}, vol.~17, pp. 2364--2377, 2022.

\bibitem{10189878}
X.~Yuan, Z.~Zhang, X.~Wang, and L.~Wu, ``Semantic-aware adversarial training
  for reliable deep hashing retrieval,'' \emph{IEEE Trans. Inf. Forensics
  Secur.}, vol.~18, pp. 4681--4694, 2023.

\bibitem{PGD}
A.~Madry, A.~Makelov, L.~Schmidt, D.~Tsipras, and A.~Vladu, ``Towards deep
  learning models resistant to adversarial attacks,'' in \emph{ICLR}, 2018.

\bibitem{singh2024revisiting}
N.~D. Singh, F.~Croce, and M.~Hein, ``Revisiting adversarial training for
  imagenet: Architectures, training and generalization across threat models,''
  \emph{NeurIPS}, vol.~36, 2024.

\bibitem{FGSMRS}
E.~Wong, L.~Rice, and J.~Z. Kolter, ``Fast is better than free: Revisiting
  adversarial training,'' in \emph{ICLR}, 2019.

\bibitem{NFGSM}
P.~de~Jorge~Aranda, A.~Bibi, R.~Volpi, A.~Sanyal, P.~Torr, G.~Rogez, and
  P.~Dokania, ``Make some noise: Reliable and efficient single-step adversarial
  training,'' \emph{NeurIPS}, vol.~35, pp. 12\,881--12\,893, 2022.

\bibitem{RFATLAP}
G.~Y. Park and S.~W. Lee, ``Reliably fast adversarial training via latent
  adversarial perturbation,'' in \emph{ICCV}, 2021, pp. 7758--7767.

\bibitem{SAT}
T.~Li, Y.~Wu, S.~Chen, K.~Fang, and X.~Huang, ``Subspace adversarial
  training,'' in \emph{CVPR}, 2022, pp. 13\,409--13\,418.

\bibitem{RAFATLB}
Y.~Zhang, G.~Zhang, P.~Khanduri, M.~Hong, S.~Chang, and S.~Liu, ``Revisiting
  and advancing fast adversarial training through the lens of bi-level
  optimization,'' in \emph{ICML}.\hskip 1em plus 0.5em minus 0.4em\relax PMLR,
  2022, pp. 26\,693--26\,712.

\bibitem{SADR}
S.~Goyal, S.~Doddapaneni, M.~M. Khapra, and B.~Ravindran, ``A survey of
  adversarial defenses and robustness in nlp,'' \emph{ACM Computing Surveys},
  vol.~55, no. 14s, pp. 1--39, 2023.

\bibitem{yue2024revisiting}
X.~Yue, M.~Ningping, Q.~Wang, and L.~Zhao, ``Revisiting adversarial robustness
  distillation from the perspective of robust fairness,'' \emph{NeurIPS},
  vol.~36, 2024.

\bibitem{TBRTBF}
H.~Xu, X.~Liu, Y.~Li, A.~Jain, and J.~Tang, ``To be robust or to be fair:
  Towards fairness in adversarial training,'' in \emph{ICML}.\hskip 1em plus
  0.5em minus 0.4em\relax PMLR, 2021, pp. 11\,492--11\,501.

\bibitem{CFA}
Z.~Wei, Y.~Wang, Y.~Guo, and Y.~Wang, ``Cfa: Class-wise calibrated fair
  adversarial training,'' in \emph{CVPR}, 2023, pp. 8193--8201.

\bibitem{DAFA}
H.~Lee, S.~Lee, H.~Jang, J.~Park, H.~Bae, and S.~Yoon, ``Dafa: Distance-aware
  fair adversarial training,'' in \emph{ICLR}, 2023.

\bibitem{ATPD}
M.~Balunovi{\'c} and M.~Vechev, ``Adversarial training and provable defenses:
  Bridging the gap,'' in \emph{ICLR}, 2020.

\bibitem{LBGAT}
J.~Cui, S.~Liu, L.~Wang, and J.~Jia, ``Learnable boundary guided adversarial
  training,'' in \emph{CVPR}, 2021, pp. 15\,721--15\,730.

\bibitem{wang2023better}
Z.~Wang, T.~Pang, C.~Du, M.~Lin, W.~Liu, and S.~Yan, ``Better diffusion models
  further improve adversarial training,'' in \emph{ICML}.\hskip 1em plus 0.5em
  minus 0.4em\relax PMLR, 2023, pp. 36\,246--36\,263.

\bibitem{BATHE}
T.~Pang, X.~Yang, Y.~Dong, K.~Xu, J.~Zhu, and H.~Su, ``Boosting adversarial
  training with hypersphere embedding,'' \emph{NeurIPS}, vol.~33, pp.
  7779--7792, 2020.

\bibitem{FATSC}
M.~Zhao, L.~Zhang, Y.~Kong, and B.~Yin, ``Fast adversarial training with smooth
  convergence,'' in \emph{ICCV}, 2023, pp. 4720--4729.

\bibitem{huang2024accelerated}
H.~Huang and Z.~Zeng, ``An accelerated approach on adaptive gradient neural
  network for solving time-dependent linear equations: A state-triggered
  perspective,'' \emph{IEEE Trans. Neural Netw. Learn. Syst.}, 2024.

\bibitem{TEEAT}
G.~Sriramanan, S.~Addepalli, A.~Baburaj \emph{et~al.}, ``Towards efficient and
  effective adversarial training,'' \emph{NeurIPS}, vol.~34, pp.
  11\,821--11\,833, 2021.

\bibitem{jia2024revisiting}
X.~Jia, Y.~Chen, X.~Mao, R.~Duan, J.~Gu, R.~Zhang, H.~Xue, Y.~Liu, and X.~Cao,
  ``Revisiting and exploring efficient fast adversarial training via law:
  Lipschitz regularization and auto weight averaging,'' \emph{IEEE Trans. Inf.
  Forensics Secur.}, 2024.

\bibitem{UCOSS}
H.~Kim, W.~Lee, and J.~Lee, ``Understanding catastrophic overfitting in
  single-step adversarial training,'' in \emph{AAAI}, vol.~35, no.~9, 2021, pp.
  8119--8127.

\bibitem{GradAlign}
M.~Andriushchenko and N.~Flammarion, ``Understanding and improving fast
  adversarial training,'' \emph{NeurIPS}, vol.~33, pp. 16\,048--16\,059, 2020.

\bibitem{EATTAE}
H.~Zheng, Z.~Zhang, J.~Gu, H.~Lee, and A.~Prakash, ``Efficient adversarial
  training with transferable adversarial examples,'' in \emph{CVPR}, 2020, pp.
  1181--1190.

\bibitem{FGSM}
I.~Goodfellow, J.~Shlens, and C.~Szegedy, ``Explaining and harnessing
  adversarial examples,'' in \emph{ICLR}, 2015.

\bibitem{SDI}
X.~Jia, Y.~Zhang, B.~Wu, J.~Wang, and X.~Cao, ``Boosting fast adversarial
  training with learnable adversarial initialization,'' \emph{IEEE Trans. Image
  Process.}, vol.~31, pp. 4417--4430, 2022.

\bibitem{PGIFGSM}
X.~Jia, Y.~Zhang, X.~Wei, B.~Wu, K.~Ma, J.~Wang, and X.~Cao, ``Prior-guided
  adversarial initialization for fast adversarial training,'' in
  \emph{ECCV}.\hskip 1em plus 0.5em minus 0.4em\relax Springer, 2022, pp.
  567--584.

\bibitem{WAT}
B.~Li and W.~Liu, ``Wat: Improve the worst-class robustness in adversarial
  training,'' in \emph{AAAI}, vol.~37, no.~12, 2023, pp. 14\,982--14\,990.

\bibitem{GAT}
G.~Sriramanan, S.~Addepalli, A.~Baburaj \emph{et~al.}, ``Guided adversarial
  attack for evaluating and enhancing adversarial defenses,'' \emph{NeurIPS},
  vol.~33, pp. 20\,297--20\,308, 2020.

\bibitem{DenseNet}
G.~Huang, Z.~Liu, L.~Van Der~Maaten, and K.~Q. Weinberger, ``Densely connected
  convolutional networks,'' in \emph{CVPR}, 2017, pp. 4700--4708.

\bibitem{IncV}
C.~Szegedy, V.~Vanhoucke, S.~Ioffe, J.~Shlens, and Z.~Wojna, ``Rethinking the
  inception architecture for computer vision,'' in \emph{CVPR}, 2016, pp.
  2818--2826.

\bibitem{SwinTrans}
Z.~Liu, Y.~Lin, Y.~Cao, H.~Hu, Y.~Wei, Z.~Zhang, S.~Lin, and B.~Guo, ``Swin
  transformer: Hierarchical vision transformer using shifted windows,'' in
  \emph{ICCV}, 2021, pp. 10\,012--10\,022.

\bibitem{pang2022robustness}
T.~Pang, M.~Lin, X.~Yang, J.~Zhu, and S.~Yan, ``Robustness and accuracy could
  be reconcilable by (proper) definition,'' in \emph{ICML}.\hskip 1em plus
  0.5em minus 0.4em\relax PMLR, 2022, pp. 17\,258--17\,277.

\bibitem{TDAT}
K.~Tong, C.~Jiang, J.~Gui, and Y.~Cao, ``Taxonomy driven fast adversarial
  training,'' in \emph{AAAI}, vol.~38, no.~6, 2024, pp. 5233--5242.

\bibitem{AEE}
Y.~Ge, Y.~Li, K.~Han, J.~Zhu, and X.~Long, ``Advancing example exploitation can
  alleviate critical challenges in adversarial training,'' in \emph{ICCV},
  2023, pp. 145--154.

\bibitem{linover}
R.~Lin, C.~Yu, B.~Han, and T.~Liu, ``On the over-memorization during natural,
  robust and catastrophic overfitting,'' in \emph{ICLR}.

\bibitem{cifar}
A.~Krizhevsky, G.~Hinton \emph{et~al.}, ``Learning multiple layers of features
  from tiny images,'' 2009.

\bibitem{imagenet}
J.~Deng, W.~Dong, R.~Socher, L.-J. Li, K.~Li, and L.~Fei-Fei, ``Imagenet: A
  large-scale hierarchical image database,'' in \emph{CVPR}, 2009, pp.
  248--255.

\bibitem{ResNet}
K.~He, X.~Zhang, S.~Ren, and J.~Sun, ``Deep residual learning for image
  recognition,'' in \emph{CVPR}, 2016, pp. 770--778.

\bibitem{LASAT}
X.~Jia, Y.~Zhang, B.~Wu, K.~Ma, J.~Wang, and X.~Cao, ``Las-at: Adversarial
  training with learnable attack strategy,'' in \emph{CVPR}, 2022, pp.
  13\,398--13\,408.

\bibitem{MART}
Y.~Wang, D.~Zou, J.~Yi, J.~Bailey, X.~Ma, and Q.~Gu, ``Improving adversarial
  robustness requires revisiting misclassified examples,'' in \emph{ICLR},
  2019.

\bibitem{MIFGSM}
Y.~Dong, F.~Liao, T.~Pang, H.~Su, J.~Zhu, X.~Hu, and J.~Li, ``Boosting
  adversarial attacks with momentum,'' in \emph{CVPR}, 2018, pp. 9185--9193.

\bibitem{CW}
N.~Carlini and D.~Wagner, ``Towards evaluating the robustness of neural
  networks,'' in \emph{IEEE Symposium on Security and Privacy (SP)}.\hskip 1em
  plus 0.5em minus 0.4em\relax IEEE, 2017, pp. 39--57.

\bibitem{AutoAttack}
F.~Croce and M.~Hein, ``Reliable evaluation of adversarial robustness with an
  ensemble of diverse parameter-free attacks,'' in \emph{ICML}.\hskip 1em plus
  0.5em minus 0.4em\relax PMLR, 2020, pp. 2206--2216.

\bibitem{torchattacks}
H.~Kim, ``Torchattacks: A pytorch repository for adversarial attacks,''
  \emph{arXiv preprint arXiv:2010.01950}, 2020.

\bibitem{PGK}
X.~Jia, Y.~Zhang, X.~Wei, B.~Wu, K.~Ma, J.~Wang, and X.~Cao, ``Improving fast
  adversarial training with prior-guided knowledge,'' \emph{IEEE Trans. Pattern
  Anal. Mach. Intell.}, 2024.

\bibitem{lossSurface}
Z.~Ge, H.~Liu, W.~Xiaosen, F.~Shang, and Y.~Liu, ``Boosting adversarial
  transferability by achieving flat local maxima,'' \emph{NeurIPS}, vol.~36,
  pp. 70\,141--70\,161, 2023.

\bibitem{WRN}
S.~Zagoruyko and N.~Komodakis, ``Wide residual networks,'' in
  \emph{BMVC}.\hskip 1em plus 0.5em minus 0.4em\relax British Machine Vision
  Association, 2016.

\end{thebibliography}

\end{document}